\documentclass[journal]{IEEEtran}

\usepackage{cite}

\usepackage{siunitx} 

\ifCLASSINFOpdf
   \usepackage[pdftex]{graphicx}
\else
\fi
\usepackage{amsmath}
\usepackage{amsfonts} 
\usepackage{array}

\usepackage{xcolor}
\mathchardef\mhyphen="2D

\newcommand{\RNum}[1]{\uppercase\expandafter{\romannumeral #1\relax}}

\ifCLASSOPTIONcompsoc
 \usepackage[caption=false,font=normalsize,labelfont=sf,textfont=sf]{subfig}
\else
 \usepackage[caption=false,font=footnotesize]{subfig}
\fi
\usepackage{mathtools}

\usepackage{lineno} 

\begin{document}
\title{Neural Network Normal Estimation and Bathymetry Reconstruction from Sidescan Sonar}

\author{Yiping Xie,
        Nils Bore
        and~John~Folkesson
\thanks{This work was partially supported by the Wallenberg AI, Autonomous Systems and Software Program (WASP) funded by the Knut and Alice Wallenberg Foundation and partially supported by Stiftelsen för Strategisk Forskning
(SSF) through the Swedish Maritime Robotics Centre (SMaRC)
(IRC15-0046). (Corresponding author: Yiping Xie.)}
\thanks{The authors are with the Robotics, Perception and Learning Lab, Royal Institute of Technology, SE-100 44, Stockholm, Sweden (e-mail: yipingx@kth.se; nbore@kth.se; johnf@kth.se).}
\thanks{Digital Object Identifier 10.1109/JOE.2022.3194899}
}

\maketitle

\begin{abstract}
Sidescan sonar intensity encodes information about the changes of surface normal of the seabed. However, other factors such as seabed geometry as well as its material composition also affect the return intensity.  One can model these intensity changes in a forward direction from the surface normals from bathymetric map and physical properties to the measured intensity or alternatively one can use an inverse model which starts from the intensities and models the surface normals.  Here we use an inverse model which leverages deep learning's ability to learn from data; a convolutional neural network is used to estimate the surface normal from the sidescan. Thus the internal properties of the seabed are only implicitly learned.  Once this information is estimated, a bathymetric map can be reconstructed through an optimization framework that also includes altimeter readings to provide a sparse depth profile as a constraint. Implicit neural representation learning was recently proposed to represent the bathymetric map in such an optimization framework. In this article, we use a neural network to represent the map and optimize it under constraints of altimeter points and estimated surface normal from sidescan. By fusing multiple observations from different angles from several sidescan lines, the estimated results are improved through optimization. We demonstrate the efficiency and scalability of the approach by reconstructing a high-quality bathymetry using sidescan data from a large sidescan survey. We compare the proposed data-driven inverse model approach of modeling a sidescan with a forward Lambertian model.  We assess the quality of each reconstruction by comparing it with data constructed from a multibeam sensor.  
\end{abstract}

\begin{IEEEkeywords}
Bathymetry reconstruction, Implicit neural representations, Neural networks, Sidescan sonar
\end{IEEEkeywords}


%
\IEEEpeerreviewmaketitle

\section{Introduction}

\IEEEPARstart{I}{n} recent years there has been increasing interest in the use of sidescan to reconstruct bathymetric maps. Bathymetric maps are usually constructed with high-end multibeam echo sounders (MBES), which are normally mounted on survey vessels or large autonomous underwater vehicles (AUVs). However, such MBES are relatively large and expensive compared to sidescan sonars, thus not suitable for smaller AUVs. Furthermore, sidescans generally have a wider swath range than multibeam and can produce images with a much higher resolution. If information about the seafloor's slope changes can be inferred from sidescan images, a low-cost and efficient method to construct high-resolution bathymetric maps would result and be of great benefit to many applications using smaller AUVs.

For sidescan intensities, we know the exact range to the seafloor but not the beam angle due to the fact that the transducer array is along a single line.  The two-way travel time of the signal gives an accurate range to the reflection, and the reflected acoustic signals also contain some information about the seafloor's slope, e.g., the changes of the intensities indicate the changes of the incidence angle \cite{folkesson20}. Reconstruction of the surface of the seafloor from 2D sidescan image intensities is a mathematically ill-posed problem. Traditionally, shape-from-shading techniques have been used to derive the slopes of seafloor with gridded bathymetric data as control points \cite{li1991improvement}, under the assumption of a Lambertian model \cite{lambertian1991}. Recently, using deep learning approaches \cite{xie2019inferring}, \cite{xie2021} we have shown very promising results by estimating depth contours from sidescan images with a neural network on large-scale sonar data. Our prior work \cite{xie2019inferring} shows that the estimation errors from a single sidescan line using a CNN are most significant as one moves further from the sensor in the across-track direction. Additional constraints from adding the altimeter readings along the trajectories of other lines were able to reduce this problem at longer ranges  \cite{xie2021}. However, the accuracy of the altimeters worsens in deeper waters, making the constraint less accurate, thus affecting the bathymetry reconstruction.

\begin{figure*}[t]
\centering
\includegraphics[width=6.5in]{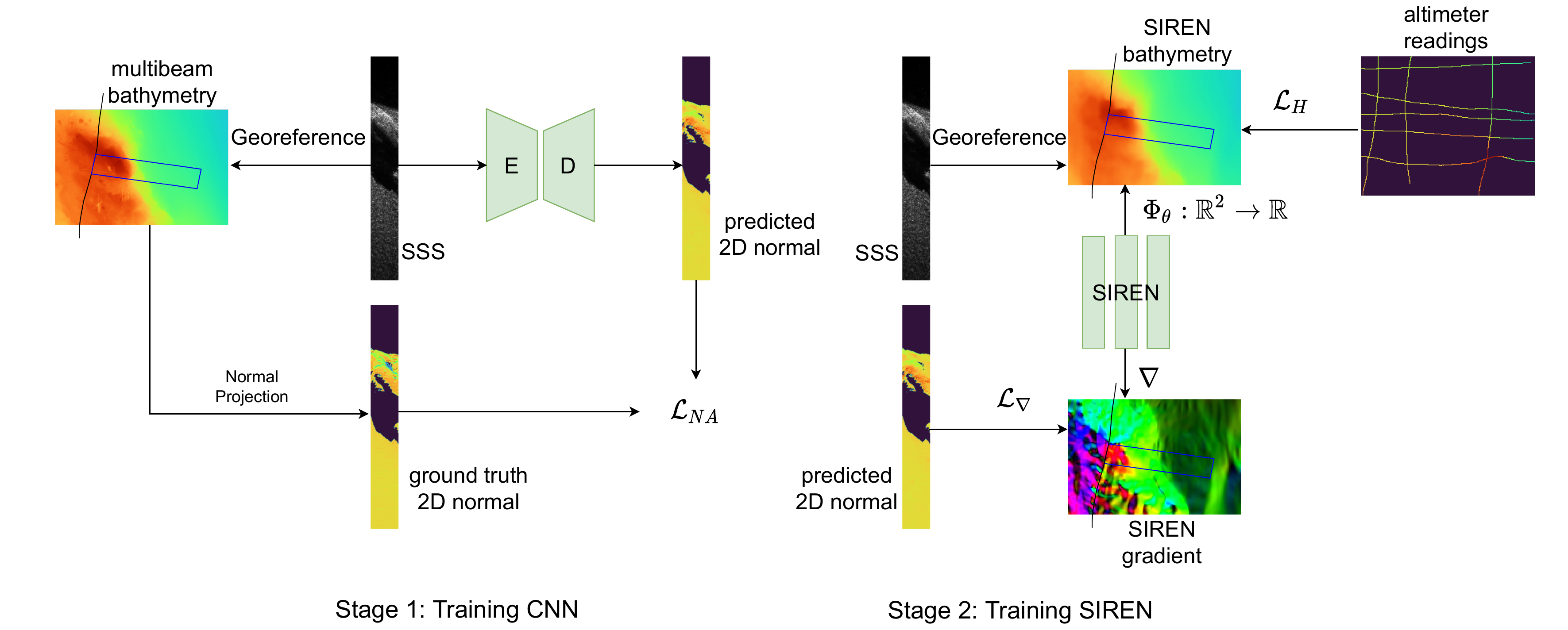}
\caption{Overview: The method has two stages. Only the first requires MBES data while the second stage can be repeated on new surveys without MBES data once the first training stage has been completed. Stage 1: Given the sidescan sonar (SSS) data and the mesh formed from MBES data, we can georeference the sidescan images to create training data for learning the inverse sensor model that estimates the surface normal from sidescan with a CNN. Such CNN (fully convolutional network with an encoder "E" and a decoder "D") is trained with normal-aware loss, which can focus both the high normal area and the low normal  area. Stage 2: We use the surface normal predicted from the only sidescan data using the now trained CNN to constrain the gradient of the SIREN MLP, i.e., the to be estimated bathymetry. At the same time, we also use the altimeter readings to constrain the SIREN directly. This is also the testing stage for the CNN, since we use the test set of the CNN to predict the surface normal and estimate the bathymetry.}
\label{fig:overview}
\end{figure*}

A typical survey has considerable overlap between lines and exploiting that to further constrain the bathymetric map is an aim of this work.  Each sidescan can only measure surface slopes in the plane with normal in the direction of the linear transducer array.  To measure slopes in other planes one must see the seabed from other viewpoints.  However, to reconstruct bathymetry from several sidescan lines, the fusion of different bathymetric estimates needs to be addressed. Previously we estimated the uncertainty of the neural network depth prediction and used that uncertainty for confidence weighting during the fusion process \cite{xie2021}. We also used a different approach, treating it as an optimization problem with a physical model to predict the sidescan intensities rather than a neural network while still reconstructing the bathymetry from multiple sidescan lines \cite{nils2021}. That approach, often referred to as \textit{inverse rendering}, comes from the field of Computer Vision and attempts to estimate reflectance, geometry, and lighting from multiple camera images in an optimization framework. Recently, more and more focus has been moved towards \textit{differentiable rendering} where the forward rendering process is differentiable so that optimizing the loss between camera images and rendered images would result in better estimates of the 3D geometry. And very recently, \textit{representation learning} has been adapted into the differentiable rendering pipeline, being able to produce state-of-the-art results where neural networks are used as representations of the 3D scene geometry \cite{srn2019}\cite{nerf2020}\cite{siren2020}. Particularly, Bore \textit{et al.} \cite{nils2021} has shown that the differentiable representations parameterized by neural networks proposed in \cite{siren2020} allow constraining the slopes of the seafloor to match the one inferred from a forward sensor model of the sidescan intensity under a Lambertian assumption.

There are several problems of using a Lambertian model to render the measured sidescan intensity. Firstly, Lambertian reflection model is based on the assumption of ideal diffuse reflection thus cannot properly approximate the non-diffuse reflection. Secondly, Lambert's reflection law, modeling the power of the received signal is proportional to the cotangent of the angle between the surface normal and incoming sound \cite{folkesson20}.  Nevertheless many authors use other dependencies, for example  Aykin \textit{et al.} \cite{aykin2013forward} and Dur{\'a} \textit{et al.}  \cite{dura2004reconstruction} model the received energy as proportional to the square cosine of the incidence angle.  Therefore we propose to replace the physical model with a convolutional neural network (CNN), namely using a data-driven approach to estimate the surface normal from sidescan intensity and subsequently constraining the seafloor slopes of the estimated bathymetry. Another motivation of using a data-driven approach is that a neural network could potentially learn the reflectivity from context, e.g., if it looks like a rock from the sidescan images, it probably reflects like a rock.

The proposed method for reconstructing bathymetry from sidescan and altimeter readings has two stages (see Fig. \ref{fig:overview}), a training stage that requires training data with some ground truth bathymetry (i.e., MBES data from the training area), and an optimization stage that can be repeated for many new survey regions and only requires the collection of sidescan data, altimeter and navigation.   The first stage is to train a CNN to estimate surface normals from sidescan waterfall images. To train the CNN, we first  need to georeference the sidescan data, that is to register the sidescan images with the mesh formed from MBES data, to create training pairs. After the CNN is trained, we enter the second stage, where we have another neural network, a Multi-layer Perceptron (MLP), to represent the bathymetry that we want to reconstruct. Such a neural network allows access to its gradient hence allowing us to supervise on the gradient of the estimated bathymetry directly. We use the sidescan and altimeter data from the test set of the trained CNN to estimate the bathymetry by using the surface normals predicted from the CNN to constrain the gradients of the bathymetry and altimeter readings to constrain the bathymetry itself.

The {\bf contribution} of this work is to combine and extend our previous work \cite{nils2021, xie2021}.  We evaluate that numerically using real survey data containing both sidescan and MBES collected simultaneously.  Thus we present a fair comparison of the two approaches of either using the forward physical sensor model versus using an inverse neural network sensor model for the surface reflection process.

\subsection{Related work}

There have been many attempts to estimate the bathymetry from sidescan, most of which are model-based. Many of the methods are inspired by the technique known as \textit{shape-from-shading}, which was first proposed by Horn \cite{horn1975obtaining} to estimate the shape of objects from shading information with camera images. Lander \& Hebert \cite{lambertian1991} and Li \& Pai \cite{li1991improvement} were among the earliest to use such technique to sidescan sonar images. One of the differences between \cite{lambertian1991} and \cite{li1991improvement}  is that in \cite{lambertian1991} they model the sonar's reflection model as a factorization of the diffuse and specular components, while in \cite{li1991improvement} they use the Lambertian model.  
Another series of model-based methods \cite{coiras2007multiresolution} \cite{coiras2009simulation} use a Lambertian model for the surface scattering process and estimate the bathymetry, the reflectance and the 
beam-pattern from sidescan at the same time, with an expectation-maximization (EM) approach using gradient descent. 

Since estimating bathymetry from sidescan is an ill-posed inverse problem, many works have introduced some boundary conditions from bathymetric data. Li \& Pai \cite{li1991improvement} proposed  improving the MBES bathymetry with high-resolution sidescan data when the bathymetric data from MBES is of too coarse a resolution to describe the detailed terrain of the seafloor. The grid-based, sparse bathymetric data is used as control points for the shape-from-shading process. Johnson \& Hebert \cite{johnson1996seafloor} suggest using the sparse bathymetric data as an initial estimate for the bathymetry, then iteratively refining the bathymetric map by minimizing the error between sidescan backscatter data and the one predicted by the bathymetry. They use smoothness regularization as a constraint. Since they phrase the fitting process as a global optimization problem, in principle, their approach can deal with multiple observations from different sidescan lines. Bore \& Folkesson \cite{nils2021} also introduce sparse altimeter readings as a constraint and phrase the bathymetry reconstruction from multiple sidescan lines as a global optimization problem, enforcing smoothness prior on the estimated bathymetry, thus increasing robustness against sensor noise or model errors. They also propose the utilization of GPUs to accelerate such an optimization.

In the last decades, a great deal of research into shape-from-shading has focused on data-driven approaches, most of them deep learning. Some recent works have used neural networks to predict the depth \cite{liu2015deep} \cite{laina2016deeper} \cite{LI2018328} \cite{attentionloss2018}, or surface normals \cite{wang2015designing} \cite{Bansal_2016_CVPR} or both \cite{eigen2015predicting} \cite{Li_2015_CVPR} \cite{Qi_2018_CVPR} from camera images and have achieved state-of-the-art results. Bathymetry reconstruction from sidescan can be seen as an analogous task to 3D reconstruction from monocular camera images. In our previous work \cite{xie2019inferring}, a method was proposed to use a CNN to estimate the depth contours of the seafloor from sidescan images with smoothness regularization. To reduce the gradual increase in depth estimation error as one moves further away from the sidescan sensor, our prior work \cite{xie2021} introduced sparse depth constraints from altimeter readings along the sonar's trajectory and reconstructed the bathymetry by fusing multiple bathymetric estimates along with the corresponding prediction uncertainty into a single map.

In the recent years, deep learning has been used on sidescan images for other tasks, such as classification \cite{dzieciuch2016non} \cite{huo2020underwater} \cite{Steiniger2020classification}, object detection \cite{einsidler2018deep} \cite{yu2021real}, bottom tracking \cite{yan2020real} \cite{qin2021bottom},  semantic segmentation \cite{rahnemoonfar2019semantic} \cite{wu2019ecnet}, sidescan generative modeling and simulation \cite{Bore2020}. 
Dzieciuch \textit{et al.} \cite{dzieciuch2016non} show that a simple CNN can be used for mine detection in sidescan sonar imagery and achieve comparable accuracy as human operators. Huo \textit{et al.} \cite{huo2020underwater} publish a sidescan classification dataset of civilian objects and propose a method to generate semi-synthetic data to handle the imbalanced training data. They show that with a pre-trained CNN model on camera images, applying deep transfer learning on sonar images can achieve high accuracy on a classification task. Steiniger \textit{et al.} \cite{Steiniger2020classification} compare several ways of dealing with the imbalanced datasets for classification tasks, among which the augmentation of the training data with Generative Adversarial Networks (GANs) \cite{goodfellow2014generative}, e.g., BAGANs \cite{mariani2018bagan} could improve the performance of the CNN classifier. 
Einsidler \textit{et al.} \cite{einsidler2018deep} adapt the state-of-the-art object detection algorithm, YOLO (You Only Look Once) \cite{redmon2016you} on sidescan images and show that deep transfer learning on a pre-trained CNN model with some fine-tuning on the real sidescan dataset could achieve competitive performance on the anomaly detection task. Yu \textit{et al.} \cite{yu2021real} adapt the attention mechanism into the object detection algorithm, namely combining Transformer module \cite{vaswani2017attention} with YOLO, and achieve high accuracy and efficiency for underwater objection detection on sidescan images. Although they claim the algorithm could run in real-time, they conduct the experiments on a desktop with a modern CPU and GPU, whose computational power outperforms most of the onboard computers of AUVs. 
Yan \textit{et al.} \cite{yan2020real} are the first to apply 1D CNN on sidescan intensity for bottom tracking. They use a sliding window on each sidescan ping to detect the bottom position. They also claim the inference time is short enough for real-time applications but as \cite{yu2021real}, their experiments are tested with a modern CPU and GPU. Similarly, Qin \textit{et al.} \cite{qin2021bottom} use a 1D CNN to detect the bottom from sidescan data. They model the bottom tracking as a binary semantic segmentation problem on 1D sidescan data and show that U-Net based architecture has a better performance when the data is noisy.
Rahnemoonfar \& Dobbs \cite{rahnemoonfar2019semantic} propose a novel CNN architecture and illustrate it outperforms the state-of-art techniques on pothole semantic segmentation of sidescan images. Wu \textit{et al.} \cite{wu2019ecnet} propose a novel CNN named ECNet to perform semantic segmentation on sidescan with much fewer parameters and much faster speed, making it possible to be applied to real-time tasks on an embedded platform, where their experimental tests were run to compute the inference time.
Bore \& Folkesson \cite{Bore2020} propose a data-driven approach to simulate realistic sidescan given a bathymetry model with a conditional GAN. In a user survey, they show that the simulated sidescan are realistic enough to be mistaken by sidescan experts.

Much of the current work on 3D reconstruction pays particular attention to
neural rendering and implicit neural representations (INR). Neural rendering aims to produce photo-realistic renderings of a 3D scene from arbitrary viewpoints, given the representation of the scene, such as geometry, semantic structure, material illumination and appearance. By integrating differentiable rendering into the training of the neural networks, neural rendering allows us to address both reconstruction and rendering during joint optimization \cite{review-neural-rendering}. There is another line of research, implicit neural representations, focusing on different representations for modeling objects in the 3D scene, sometimes also referred to as coordinate-based representations. Other than explicitly describing scenes as voxels, point clouds or meshes, these representations such as Occupancy Networks \cite{mescheder2019occupancy}, Deep Signed Distance Function (DeepSDF) \cite{park2019deepsdf}, Scene Representation Networks (SRNs) \cite{srn2019}, Neural Radiance Fields (NeRF) \cite{nerf2020}, and Sinusoidal Representation Networks (SIRENs) \cite{siren2020} are continuous, differentiable and parameterized by MLPs. Due to the continuous nature of implicit neural representations, it can be more memory efficient than discrete representations since the model size is independent of the spatial resolution; however, the ability to model the fine detail is no longer limited by the discrete grid resolution but by the capacity of the neural networks \cite{siren2020}. Being differentiable means that gradients can be computed using automatic differentiation, enabling us to supervise the gradients directly. To better represent the fine detail in the signals, which contains high-frequency variation in the data, Mildenhall \textit{et al.} \cite{nerf2020} propose to use a sinusoidal function for positional encoding, mapping the inputs to a higher dimensional space. Subsequently, Sitzmann \textit{et al.} \cite{siren2020} show that directly using the sine function as a periodic activation function in the MLPs, namely SIRENs, not only outperforms the positional encoding strategy proposed in \cite{nerf2020}, but also allows constraints on the derivatives of a SIREN to reconstruct signals since the derivative of a SIREN is merely a phase-shifted SIREN. In \cite{siren2020} they show the great potential of SIRENs on reconstructing scenes with high quality and fidelity.
Bore \& Folkesson \cite{nils2021} show that combining a physical Lambertian model with implicit neural representations can be used for estimating a self-consistent bathymetric map from only sidescan images and altimeter readings from a large survey.

Another line of research that is closely related to this work is depth and normal estimation from monocular camera images with CNNs. Similar to the empirical discovery \cite{laina2016deeper}\cite{LI2018328}\cite{attentionloss2018} that there exists a long-tailed distribution in the histogram of per-pixel depth values in both indoor and outdoor datasets, the depth or surface normal in our existing datasets also exhibits a severe long-tailed distribution. After all, the seafloor often appears featureless. The vanilla $\ell_1$ or $\ell_2$ loss function generally used in such pixel-level regression problems will consequently focus more on the "head" regions. In the monocular depth estimation task, it will result in poorer performance on distant depth regions, while in the depth estimation from sidescan task, it would perform poorly on the "tail" regions where the seafloor has more features such as boulders and rocks. A common solution to deal with the imbalanced depth distribution is to use a depth-aware loss\cite{laina2016deeper}\cite{attentionloss2018}, which weights the errors of pixels according to the corresponding depth, forcing the neural networks to pay more attention to the "tail" regions during the training.

\subsection{Summary of the Proposed Method and the Difference From Prior Work}

Our work is based on our optimization framework in \cite{nils2021} but here, we replace the physical model with a  neural net similar to our method in \cite{xie2021}. The physical model is an approximate forward sensor model while the neural net is a data-driven inverse sensor model.   The two are likely to produce different results.  The physical model has the advantage of not needing any MBES data and can form estimates solely from sidescan.  It however must make some rather strong assumptions on the reflection process and the seabed properties.    The neural net, on the other hand, can learn the seabed properties and other factors implicitly and so can in theory be more accurate in some situations.  However, the inverse sensor model needs MBES data collected along with sidescan in order to be trained.  Once trained it can be used without MBES data.    The reconstructed bathymetric map optimization is then done as in \cite{nils2021}, for both models.

The method can be summarized as using a model of the sidescan detected reflection intensity as a function of the seafloor bathymetry.  The seafloor bathymetry map is generated by a SIREN \cite{siren2020} neural network that takes x, y seafloor coordinates as input and outputs the z coordinate.    These two models are then combined with an actual sidescan to form a loss that can be used to train the bathymetry map to better match the sidescan.  Additional constraints are added to the loss from the altimeter readings to help constrain the result. 

In \cite{nils2021} the loss is formed by simply predicting a sidescan from the models and then differencing that with the actual sidescan.  In this work, we changed that slightly to estimating a seafloor slope (normal) from the aforementioned SIREN bathymetry map and then comparing that to a seafloor normal estimated from the sidescan by a second neural network.  

\section{Method}\label{par:method}
We begin by illustrating the operation and characterization of sidescan sonar and introducing the notations. Subsequently, we describe how to use a CNN to estimate surface normals from sidescan and the loss function to tackle the long-tailed distribution in the dataset. Finally, we describe how to optimize the bathymetric representation given sidescan and the estimated normals.
\subsection{Sidescan Sonar Formation}
\begin{figure}[t]
\centering
\includegraphics[width=3.5in]{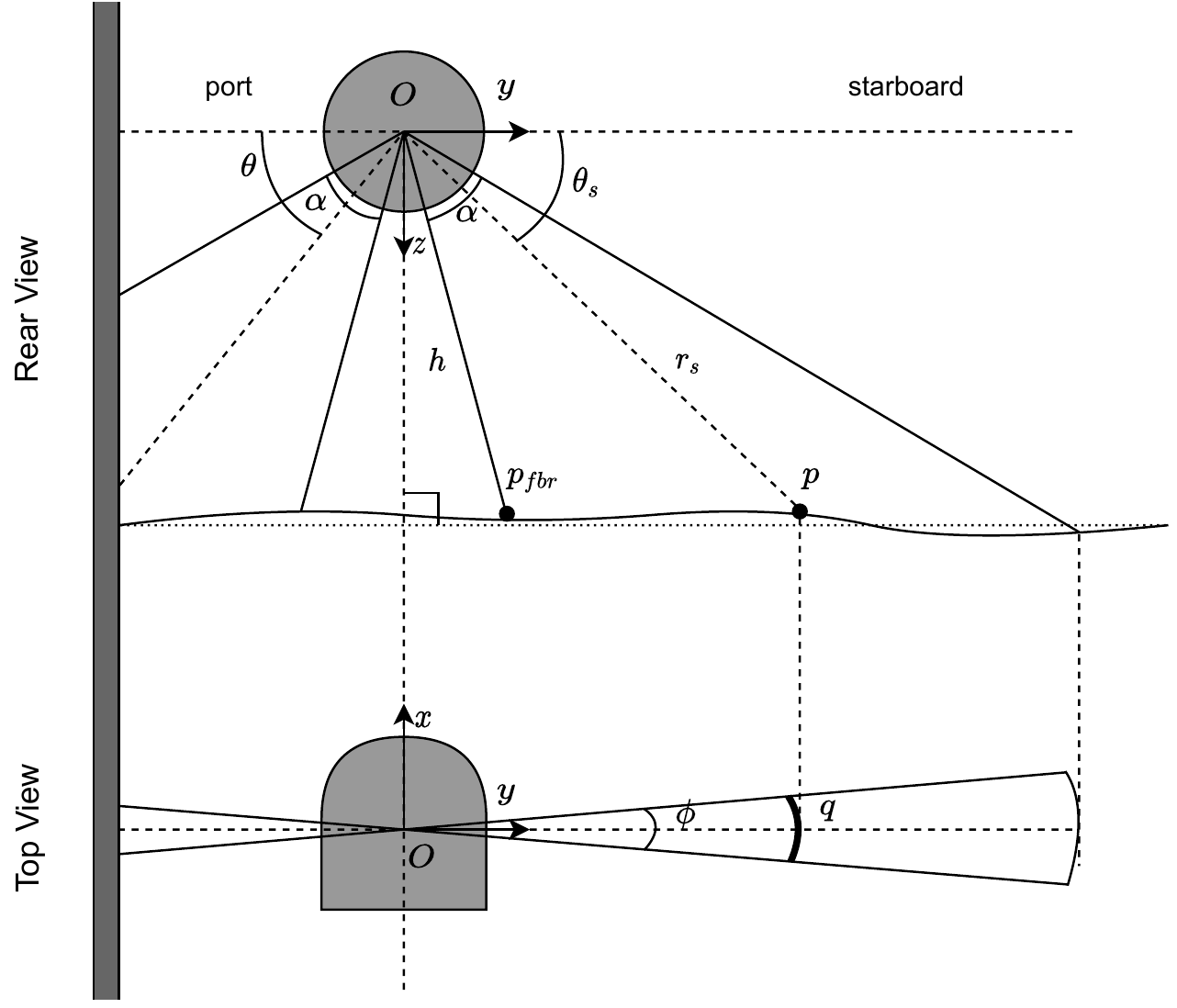}
\caption{Sidescan sonar formation in Forward Lateral Down (FLD) frame. The sidescan sonar at altitude $h$ has two heads, port and starboard, symmetrically placed to each side with a fixed angle $\theta$, angle between the horizontal axis and beam center, often referred to as \textit{tilt angle}.
The vertical beam width $\alpha$, often referred to as sensor opening in the $y \mhyphen z$ plane, is usually $40\text{-}60^\circ$, and the horizontal beam width $\phi$, often referred to as sensor opening in the $x \mhyphen y$ plane, is usually around $0.1^\circ$ \cite{blondel2010handbook}. Let $p$ be a point in the ensonified region on the bathymetric surface  with polar coordinate expressed in its slant range $r_s$ and its grazing angle $\theta_s$. Due to the horizontal beam width $\phi$, the exact point position of $p$ in the $x \mhyphen y$ plane is ambiguous over the arc $q$; however, the assumption is usually that this fact can be neglected since $\phi$ is very small. After sound travels through the water column, the first meaningful return coming from the seabed is called first bottom return, $p_{fbr}$, the detection of which can be used to determine the sonar's altitude when altimeter data is not available.}
\label{fig:sss-formation}
\end{figure}

Sidescan sonar's geometry can be described as Fig. \ref{fig:sss-formation}. Sidescan emits a beam with horizontal beam width $\phi$ that is very narrow along the direction of travel. However, sidescan has a very wide beam pattern with vertical beam width $\alpha$ in the azimuth direction, which is perpendicular to the travel direction. After the pulse of sound waves, also referred to as \textit{ping}, is sent, the sensor records the received signals comprising backscatter intensities at fixed intervals of time. These recorded intensities are arranged in a vector, which corresponds to a row in the sidescan image. Each item in this vector, often referred to as a \textit{bin}, stores the reflected intensity received from a certain distance away from the sonar array. In theory, there can be multiple reflections corresponding to the same distance, but usually, the strongest reflection comes from the surface of the seafloor. Moreover it is commonly assumed that there is only one planar surface for one bin that creates the reflection. Let point $p$ be the one responsible for the reflected echo, see Fig. \ref{fig:sss-formation}, the Time of Flight (TOF) corresponding to that bin can be used together with the Sound Velocity Profile (SVP) to determine the \textit{slant range} $r_s$ of point $p$. However, the other factor to determine the position of $p$, the \textit{grazing angle} $\theta_s$ is unknown, resulting in point $p$ could be anywhere within the interval of $[\theta-\frac{\alpha}{2},\theta+\frac{\alpha}{2}]$, depending on the surface geometry. 

\subsection{Sidescan Georeferencing}\label{sec:sidescan-draping}
\begin{figure}[t]
\centering
\includegraphics[width=3.5in]{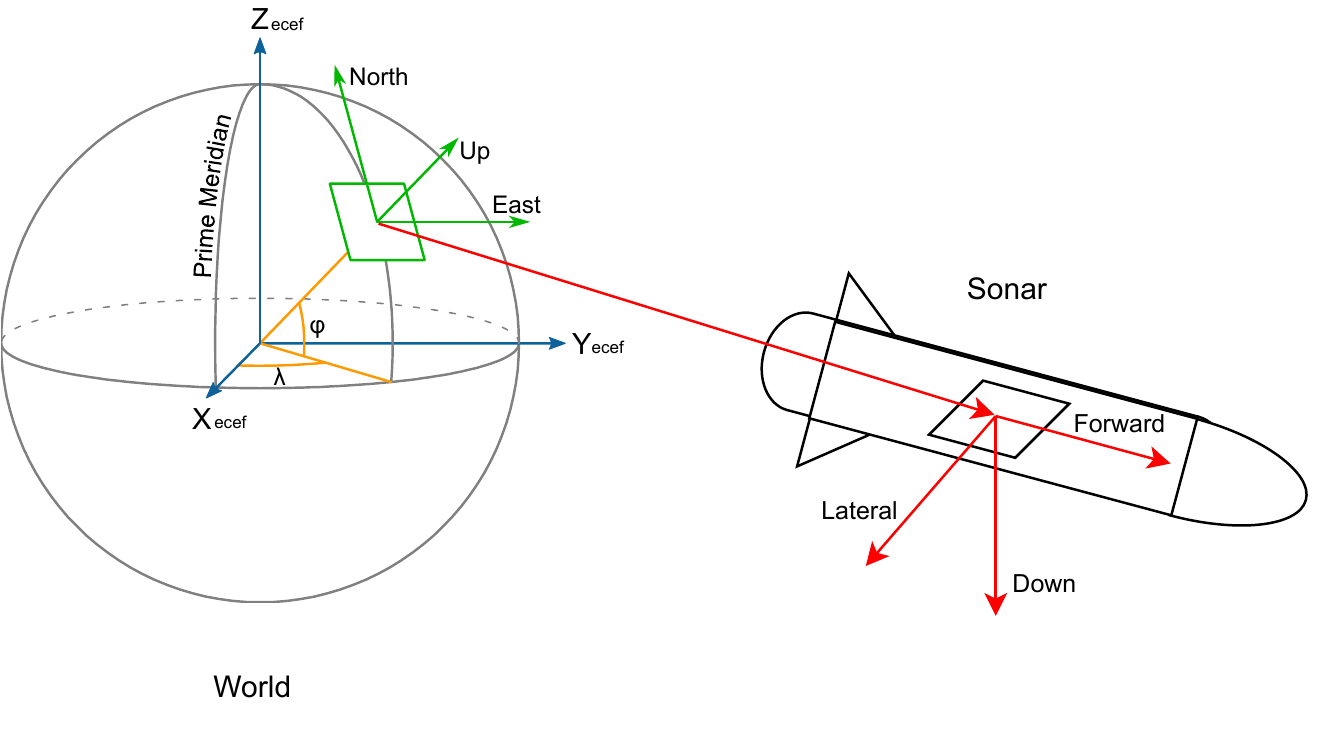}
\caption{The world and sonar frame notation. We use Earth North Up (ENU) for the world coordinates and FLD frame for the sonar frame. $\lambda$ is longitude and $\phi$ is latitude.}
\label{fig:enu-fld}
\end{figure}

To generate the training set with ground truth for training a neural network to estimate the surface normal from sidescan image, we need to associate sidescan intensity for each bin to its georeferenced coordinates on a bathymetric mesh, a process we often refer to as \textit{sidescan draping} \cite{Bore2020}. Such bathymetric mesh $\hat{\mathcal{M}}$ is formed based on multibeam data. For a sidescan waterfall image, let $I^{k,i}$ denote the returned reflected intensity corresponding to ping number $k$ and bin $i$ and $r_s^{k,i}$ denote the corresponding slant range. For simplicity, we assume an isovelocity SVP. Given the sensor position $\mathbf{s}^k \in \mathbb{R}^3$ for the current ping and the rotation matrix $\mathcal{R}^k\in SO(3)$ defining the transformation from the sonar frame to world frame (see Fig. \ref{fig:enu-fld}), we can apply ray tracing to find the intersection of each sidescan arc and the mesh. Once we have found the intersection, i.e., we have associated $I^{k,i}$ to its georeferenced coordinates $\mathbf{p}^{k,i} \in \mathbb{R}^3 $ on the mesh, we also can associate $I^{k,i}$ to the surface normal $\mathbf{N}(\mathbf{p}^{k,i}) \in \mathbb{R}^3$ with respect to the mesh at point $\mathbf{p}^{k,i}=(p_x,p_y,p_z)$. The normal is defined by the following, given the two gradient components $\nabla _x, \nabla _y$:
\begin{equation}\label{eq:3d-normal-global-frame}
    \mathbf{N}(\mathbf{p}^{k,i}) = [-\nabla _x(p_x,p_y), -\nabla _y(p_x,p_y), 1]^{T}.
\end{equation}
Note that $\mathbf{p}^{k,i}$ and $\mathbf{N}(\mathbf{p}^{k,i})$ are in the world coordinates and the normal $\mathbf{N}_s(\mathbf{p}^{k,i})$ in sonar frame (Fig.  \ref{fig:sss-formation}) can be calculated as:
\begin{equation}\label{eq:3d-normal-sonar-frame}
    \mathbf{N}_s(\mathbf{p}^{k,i}) = {\mathcal{R}^k}^T \mathbf{N}(\mathbf{p}^{k,i})=[n_{s,x}, n_{s,y}, n_{s,z}]^{T},
\end{equation}  
given the rotation matrix $\mathcal{R}^k$.

Note that it is the 2D normal projected to the $y \mhyphen z$ plane, i.e., $\mathbf{N}_{s,2d}(\mathbf{p}^{k,i})=[n_{s,y}, n_{s,z}]^T$, that affects the diffuse reflection and the reflected intensity \cite{folkesson20}, which is the reason that we propose to train a neural network to estimate the 2D unit normal vector 
\begin{equation}\label{eq:2d-normal-sonar-frame}
    \mathbf{\hat{N}}_{s,2d}(\mathbf{p}^{k,i})=[\frac{n_{s,y}}{\sqrt{n_{s,y}^2 + n_{s,z}^2}}, \frac{n_{s,z}}{\sqrt{n_{s,y}^2 + n_{s,z}^2}}]^T
\end{equation}
instead of estimating the full 3D normal $\mathbf{N}(\mathbf{p}^{k,i})$, from sidescan images.

\subsection{Estimating Normal from Sidescan}

A sidescan waterfall image corresponding to one survey line in our dataset usually contains around 2000 pings and 6000 bins per ping for each head (port and starboard), meaning that the raw waterfall image is too large to fit into the GPU memory in a practical manner. We divide each side of the waterfall image into smaller windows containing 64 pings (height $H=64$) and width $W=512$ downsampled from around 6000 bins. The selection of $H$ and $W$ is chosen to fit the CNN and at the same time ensure the sidescan’s across-track resolution high enough ($\sim 0.1 m$) to utilize this sidescan's advantage. 
\begin{figure}[t]
\centering
\includegraphics[width=3.5in]{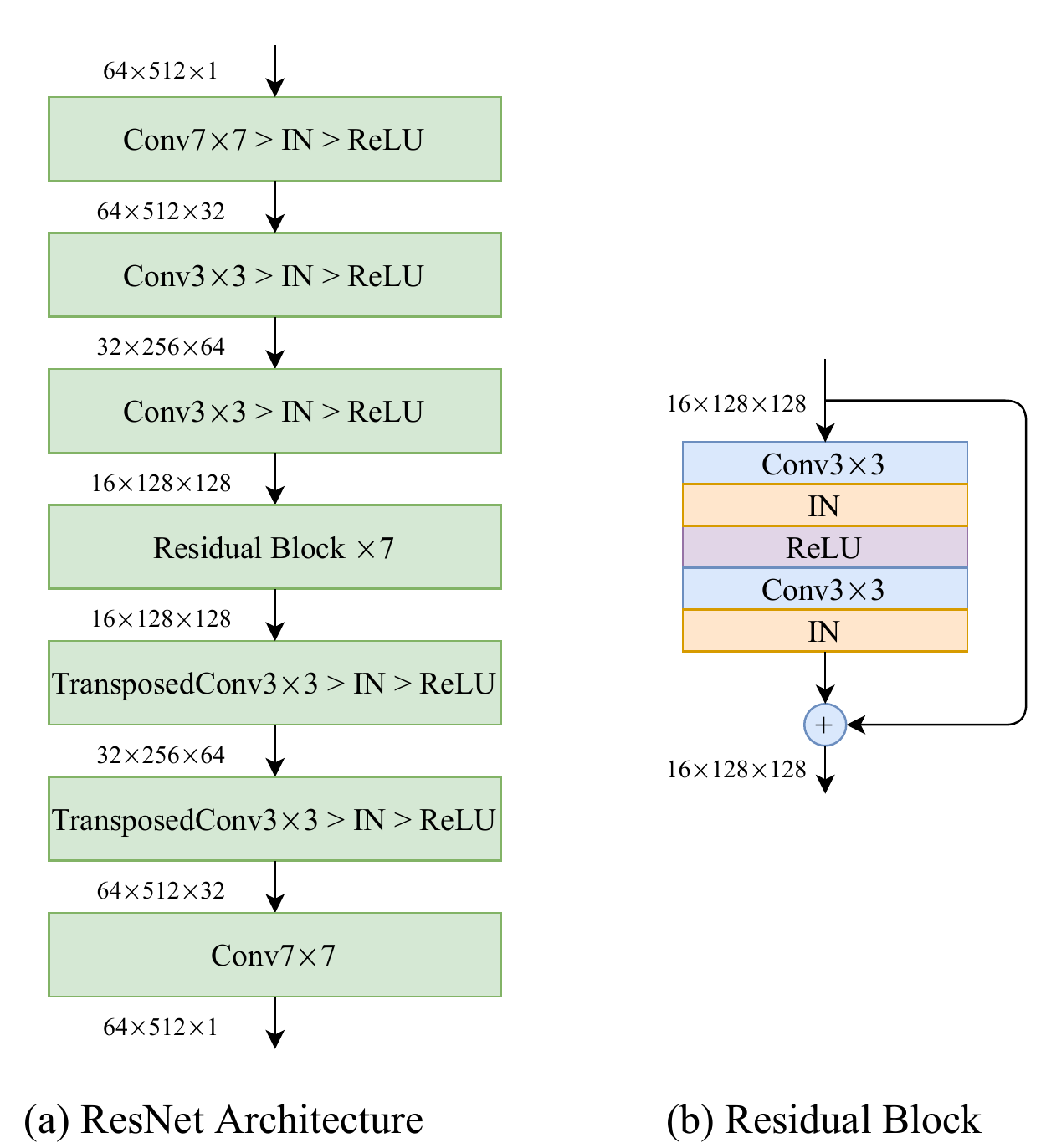}
\caption{CNN architecture, based on our prior work \cite{xie2019inferring, xie2021}, consisting of  an encoder and decoder structure with 7 residual blocks in the middle.}
\label{fig:Simplified-ResNet-sss2ndy}
\end{figure}
Similar to many monocular depth/normal estimation methods with CNNs, we formulate the task of normal prediction from sidescan intensity as a problem of learning a mapping $F: \mathcal{I} \to \mathcal{N}$ from the intensity space $\mathcal{I}$ to the output normal space $\mathcal{N}$. Since the 2D normal $\mathbf{\hat{N}}_{s,2d}(\mathbf{p}^{k,i})$ is a unit vector, we only need to estimate one of the two components, in our case the $y$ component $\hat{n}_{s,y} = \frac{n_{s,y}}{\sqrt{n_{s,y}^2 + n_{s,z}^2}}$. Hence 
the CNN model (Fig. \ref{fig:Simplified-ResNet-sss2ndy}) for learning the mapping $F$ takes the sidescan intensity window as input and outputs the corresponding $\hat{n}_{s,y}$ for every valid pixel (excluding the nadir area and shadows). Shadows are masked out as invalid data since the CNN lacks the ability to model the shadows in the sidescan images. The CNN architecture is based on our prior work \cite{xie2019inferring}\cite{xie2021}, consisting of two down-sampling layers, followed by seven residual blocks, followed by two up-sampling layers and one convolution in the end. For the normalization layer, we use Instance Normalization (IN), and for the activation functions, we use Rectified Linear Unit (ReLU).

\begin{figure}[t]
\centering
\includegraphics[width=3.5in]{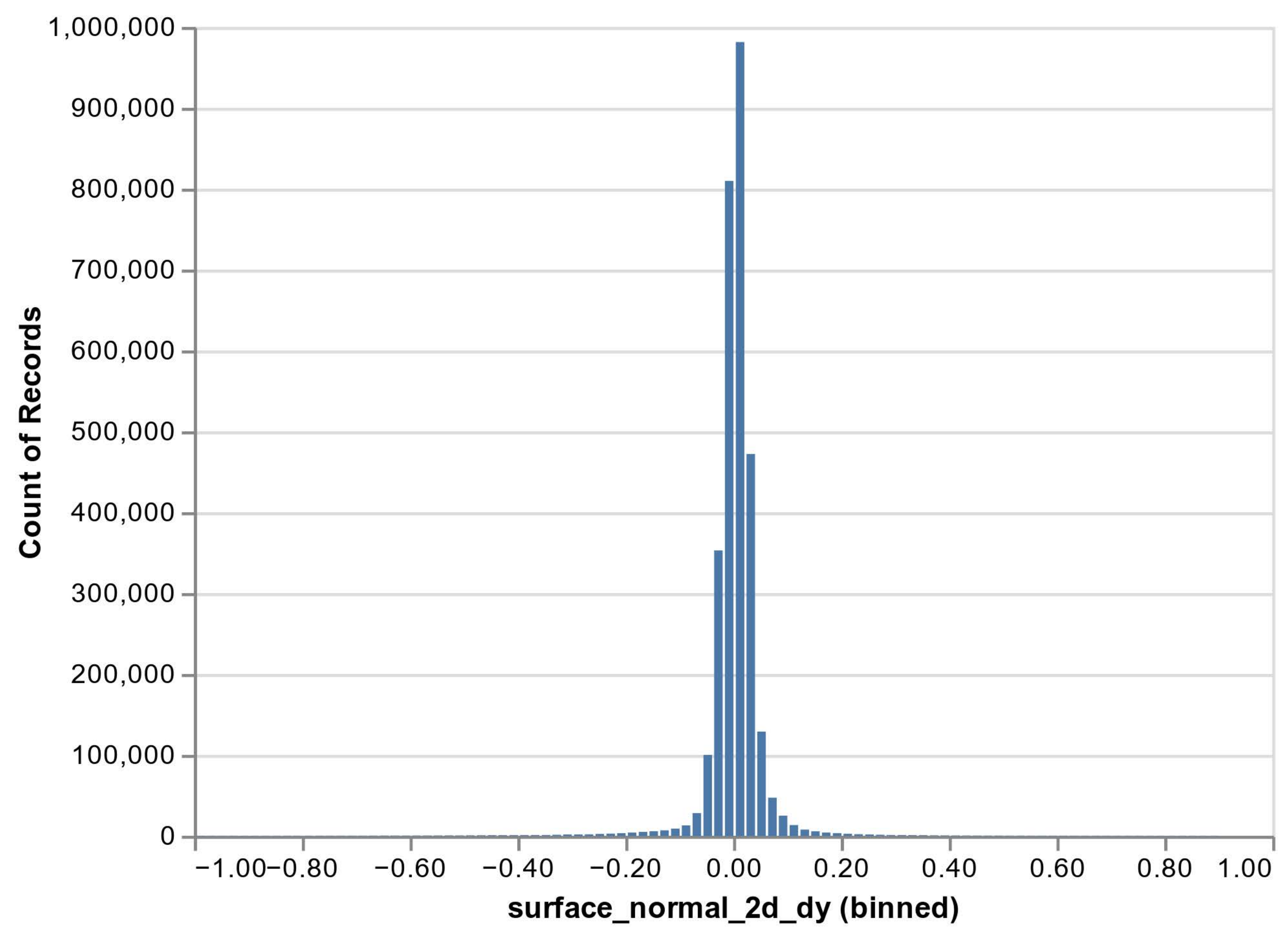}
\caption{Long-tailed distribution appeared in our data due to the nature of the seafloor. Here we show the y-component of the surface normal in sonar frame, where it is clear that the vast majority values are close to zero, indicating the seafloor is rather flat or only slowly sloping. Such distribution makes the neural network difficult to learn at areas with surface normals far from zero, e.g. hills and rocks, without explicitly addressing the long-tailed issue. }
\label{fig:long-tail-distribution}
\end{figure}
We observed that there exists an extremely long-tailed distribution in our dataset (see Fig. \ref{fig:long-tail-distribution}), which is a property of the nature that most of the seabed are relatively flat or only slightly sloping. Note that $\hat{n}_{s,y}$, i.e., the x-axis in Fig. \ref{fig:long-tail-distribution}, majorly concentrates on values between $-0.2$ and $0.2$, corresponding to $\sim \pm 10^\circ$ sloping along the lateral direction in the sonar frame. Regular $\ell_1$ or $\ell_2$ loss functions, which are commonly used in such pixel-wise regression tasks, treat all pixels equally, thus allowing easy samples with smaller absolute values to dominate the training. This will lead the network to tend to perform well only on regions with normal $\hat{n}_{s,y}$ with smaller absolute values. However, features such as rocks, boulders, ridges and hills on the seabed with large absolute normal values are of great interest. So this imbalanced problem needs to be addressed properly. Inspired by \cite{attentionloss2018}, we use a normal-aware loss function to guide the network to pay more attention to the regions with normal of large absolute values to reduce the distribution bias during the training process. The normal-aware loss is formulated as:
\begin{equation}\label{eq:normal-aware-loss}
    \mathcal{L}_{NA} = \frac{1}{\left\vert\mathcal{N}^{k,i}\right\vert}\sum_{\hat{n}_{s,y} \in \mathcal{N}^{k,i}} (\alpha_N + \lambda_N) \cdot \ell(\hat{n}_{s,y}, \hat{n}_{s,y}^{GT}),
\end{equation}
where $\mathcal{N}^{k,i}$ is the set of normal of all valid points excluding the nadir area, shadows and missing data. $\hat{n}_{s,y}$ and $\hat{n}_{s,y}^{GT}$ are the predicted normal value and ground truth value respectively. $\ell (\cdot)$ is a distance metric, $\ell_1, \ell_2$, and so on. In our case we choose to use smooth $\ell_1$ loss \cite{Girshick_2015_ICCV} similar in \cite{attentionloss2018}. $\alpha_N$ is the normal-aware attention term that guides the networks to pay more attention to the hard samples, which is positively correlated to the absolute value of the normals. We set $\alpha_N=\left\vert\hat{n}_{s,y}^{GT}\right\vert$ during the training. $\lambda_N$ is a regularization term formulated similar to \cite{attentionloss2018}: 
\begin{equation}\label{eq:alpha_n-normal-aware-loss}
    \lambda_N=1-\frac{\min(\hat{n}_{s,y}+1, \hat{n}_{s,y}^{GT}+1)}{\max(\hat{n}_{s,y}+1, \hat{n}_{s,y}^{GT}+1)},
\end{equation}
which approaches 0 when the network's prediction is close to the ground truth and approaches 1 when the network's prediction is not very accurate. Therefore, even for regions with small normal absolute values, when they are not accurately predicted, the gradients can still be back-propagated. As a result, the normal-aware loss can focus more on hard samples while preserving the attention on easy samples.

\subsection{Implicit Neural Representations}
Instead of representing the bathymetry with explicit methods (meshes or grids), we use a function $\Phi_\theta$, parameterized by a fully connected neural network with parameters $\theta$ to represent the bathymetry. The fully connected neural network is a variant of MLP with sinusoidal activation functions, known as SIREN\cite{siren2020}, mapping 2D spatial coordinates to the corresponding seafloor height, $\Phi_\theta: \mathbb{R}^2 \to \mathbb{R}$, the same as in our previous work \cite{nils2021}. Note that the representation is continuous, differentiable and capable of producing high-quality derivatives with respect to the 2D spatial coordinates, allowing us to supervise the derivatives during training. In addition to constraining the derivatives of the bathymetry, we also need some boundary conditions which come from the altimeter readings. Here we assume to have access to high-quality navigation data. The loss function of the SIREN has two parts:
\begin{equation}\label{eq:siren-loss}
    \mathcal{L} = \mathcal{L}_{\nabla} +  w_H \mathcal{L}_H,
\end{equation}
where $w_H$ is the weight for $\mathcal{L}_H$. The $\mathcal{L}_H$ loss, the same as in \cite{nils2021}, is simply trying to minimize the difference between the altimeter readings $p_z^{GT}$ and the height at ($p_x,p_y$) from the SIREN. If we take the following as the signed vertical distance:
\begin{equation}\label{eq:signed-vertical-distance}
    \Delta^{\theta}(\mathbf{p}) = \Phi_\theta(p_x,p_y) - p_z,
\end{equation}
the averaged height loss is given by:
\begin{equation}\label{eq:siren-height-loss}
\mathcal{L}_H = \frac{1}{\left\vert\mathcal{P}\right\vert} \sum_{\mathbf{p}_j^{GT} \in \mathcal{P}} \left\Vert \Delta^{\theta}(\mathbf{p}^{GT}_j) \right\Vert,
\end{equation}
where $\mathcal{P} \subset \mathbb{R}^3 $ is the set of points on the seafloor directly below the AUV.

The second part of the loss, $\mathcal{L}_{\nabla}$, aims to directly constrain on the projected 2D normal in the sonar  frame for each sidescan ping $k$ and each bin $i$:
\begin{equation}\label{eq:siren-normal-loss}
\mathcal{L}_{\nabla} = \frac{1}{\left\vert \mathcal{N}^{k,i} \right\vert} \sum_{\mathbf{\hat{N}}_{s,2d} \in \mathcal{N}^{k,i}} \left\Vert \mathbf{\hat{N}}_{s,2d}^\theta - \mathbf{\hat{N}}_{s,2d} \right\Vert,
\end{equation}
where $\mathcal{N}^{k,i}$ is the set of normal of all valid points, $\mathbf{\hat{N}}_{s,2d}$ is the estimated 2D normal in sonar  frame from sidescan intensity from CNN and  $\mathbf{\hat{N}}_{s,2d}^\theta$ is the one calculated based on bathymetry representation $\Phi_\theta$. The calculation of $\mathbf{\hat{N}}_{s,2d}^\theta$ is the same as Equation \ref{eq:3d-normal-global-frame}, \ref{eq:3d-normal-sonar-frame} and \ref{eq:2d-normal-sonar-frame} but only with the SIREN as the bathymetry. 
To calculate $\mathbf{\hat{N}}_{s,2d}^\theta$, we need to compute for each bin where the seafloor intersects the corresponding sound volume. 
The horizontal beam width $\phi$ of a sidescan is very small (see Fig. \ref{fig:sss-formation}) and usually can be neglected. Under such assumption, together with the assumption of isovelocity SVP, the reflected volume of sound lies on a thin spherical arc at a fixed distance away from the sonar. We refer to these arcs as \textit{isotemporal curves} and they are parameterized by the grazing angle $\theta_s$ for a certain slant range $r^{k,i}$:
\begin{equation}\label{eq:isotemporal-curve}
    \mathbf{p}^{k,i}(\theta_s) = \mathbf{s}^k + r^{k,i} \mathcal{R}^k [0, \cos(\theta_s), -\sin(\theta_s)]^T,
\end{equation}
where $\mathbf{s}^k$ is the sonar sensor origin, $r^{k,i}$ is the slant range and $\mathcal{R}^k$ is the rotation matrix (see Fig. \ref{fig:sss-formation} and \ref{fig:enu-fld}).  

Similarly as in our prior work \cite{nils2021}, we assume the signed vertical distance function  $\Delta^{\theta}(\mathbf{p}^{k,i}(\theta^n_s))$ only has one zero-crossing within the angle interval $[\theta-\frac{\alpha}{2},\theta+\frac{\alpha}{2}]$ (see Fig. \ref{fig:sss-formation}). Then We use gradient descent to find the crossing by starting from an initial angle $\theta^0=\theta$ and updating it iteratively:

\begin{equation}\label{eq:gd-find-crossing}
    \theta^{n+1}_s = \theta^n_s - \frac{\lambda}{r^{k,i}} \frac{d}{d \theta_s} (\Delta^{\theta}(\mathbf{p}^{k,i}(\theta^n_s)))^2,
\end{equation}
where $\lambda$ is the updating step size. Division by the slant range $r^{k,i}$, following the empirical experience in \cite{nils2021}, ensures that the step is uniform across different isotemporal curves. A fixed number of updating steps are performed and if the found angle $\theta^{*}_s \in [\theta-\frac{\alpha}{2},\theta+\frac{\alpha}{2}]$, we consider the seafloor intersection to be found. If the found angle $\theta^{*}_s$ is outside of the beam, we mark it as invalid and exclude it from the loss computation.

Once the seafloor intersection is found, the corresponding 2D projected normal in sonar frame estimated from the CNN is treated as the ground truth for the bathymetry $\Phi_\theta$. Since the process of computing 2D projected normal from $\Phi_\theta$ is differentiable, constraining the 2D projected normal would result in optimizing $\Phi_\theta$, under the constraints from the altimeter readings as the boundary conditions. 

\section{Experiments}\label{par:experiments}

\subsection{Dataset}

\begin{figure*}[t]
\centering
\includegraphics[width=6.5in]{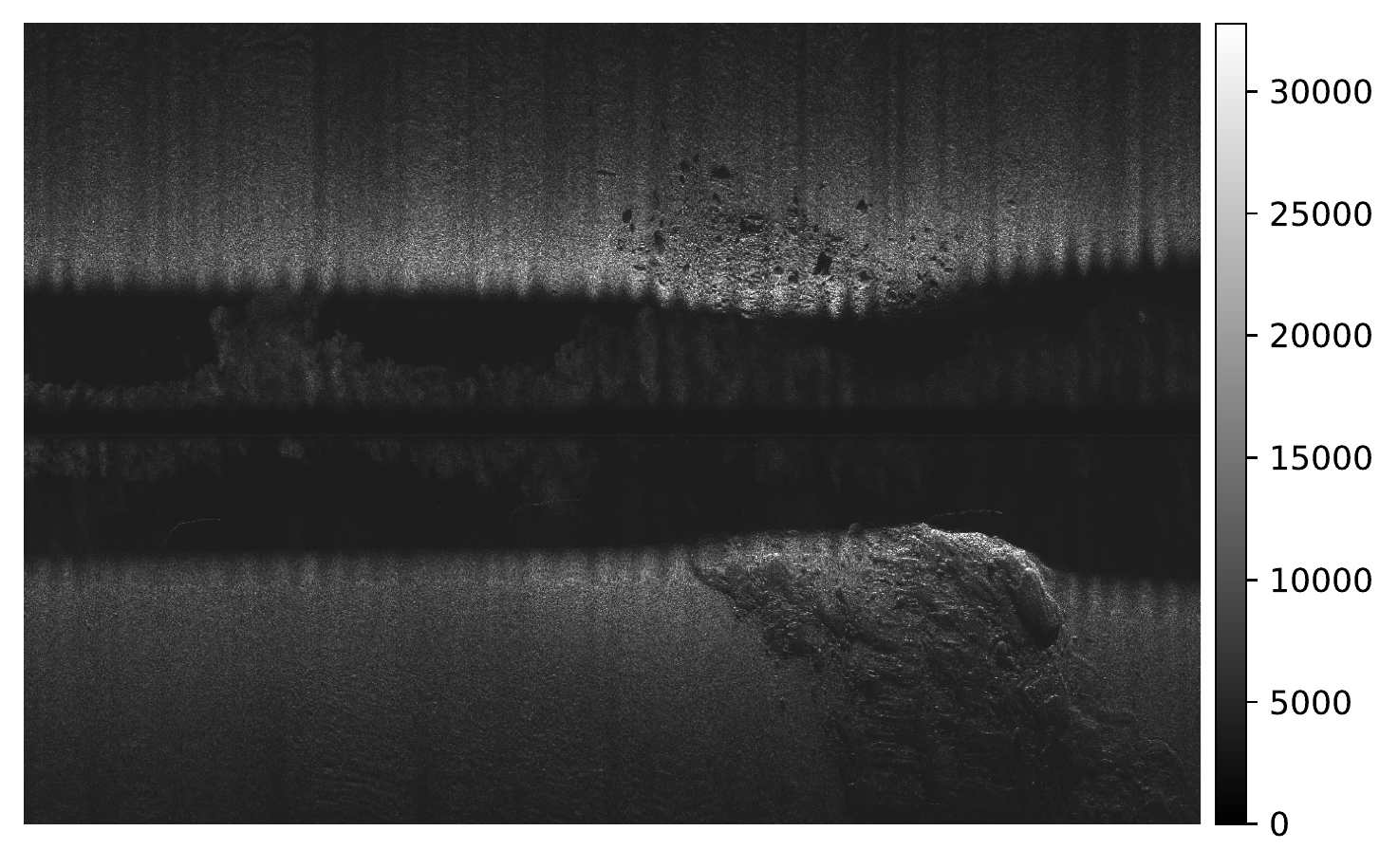} 
\caption{Downsampled sidescan image. Example of a downsampled sidescan image showing consecutive swaths. Each column corresponds to one ping, i.e., one single swath, having 512 bins downsampled from $\sim 6000$ bins in the raw image. The echo intensity is recorded and mapped to images such that black represents no echo and white denotes maximum echo intensity.}
\label{fig:sss-waterfall}
\end{figure*}

The datasets (see Table \ref{tab:datasets}) used in this paper to evaluate our method are collected with an MMT surface vessel equipped with a hull mounted sidescan Edgetech 4200MP to ensure the sensor position with high accuracy and a RTK GPS to ensure high accuracy navigation. Dataset \RNum{1} was collected from seawater and Dataset \RNum{2} was collected from freshwater. Our datasets also contain high-resolution multibeam bathymetry collected with a Reson 7125 that is taken as the ground truth for training and evaluation. There are 57 sidescan survey lines (an example shown as Fig. \ref{fig:sss-waterfall}) in total in Dataset \RNum{1}, of which 45 lines are used to train the CNN for estimating normals from sidescan, and 12 lines are used to train a SIREN bathymetry model afterward. Similar to \cite{nils2021}, we use the multibeam bathymetry that is used for evaluation and sonar's depth computed from the pressure sensor to simulate the altimeter sparse bathymetric readings so that any artifacts, bias and noise which would be introduced with a separate altimeter sensor are avoided during the evaluation of the proposed method. We use the whole of Dataset \RNum{2} including 36 sidescan survey lines to train the SIREN bathymetry to test the generalization ability of the CNN since the CNN has not been trained on any data from  Dataset \RNum{2}.

\begin{table}[t]
\renewcommand{\arraystretch}{1.5}
\caption{Datasets Details}
\label{tab:datasets}
\centering
\begin{tabular}{c||c|c}
\hline
 & Dataset \RNum{1} & Dataset \RNum{2}\\
  & \multicolumn{1}{c|}{Training Testing} & Testing \\
\hline
Survey lines & \multicolumn{1}{c|}{45 \;\;\;\;\;\;\;\; 12 } & 36\\
\hline
Bathymetry Resolution & 0.5 m & 0.5 m \\
\hline
Max Depth & 25.07 m & 13.66 m\\
\hline
Min Depth & 9.03 m & 10.45 m\\
\hline
Survey Area & $\sim 350$ m $\times \; 300$ m & $\sim 400$ m $\times \; 450$ m\\
\hline
Water Type & Salt & Fresh\\
\hline
Sidescan Range & $\sim$50 m & $\sim$50 m\\
\hline
Sidescan Frequency & 850 kHz & 850 kHz\\
\hline
\end{tabular}
\end{table}

\subsection{Training CNN}
We generate training data with the method described in Section \ref{sec:sidescan-draping}, associating every pixel in sidescan waterfall images to its corresponding 3D point on the multibeam mesh. For each valid point, we calculate the 3D surface normals from the multibeam mesh and transform and project them to the sonar frame. For either side of the sidescan waterfall image, we divide it into smaller windows with 64 pings and 512 bins (see Fig. \ref{fig:sss-gt-pre}). To generate more training data, we augment the dataset by flipping the windows in the along-track direction to simulate the sonar heading exactly the opposite direction and seeing the same place from the other side of the sonar. We also allow the consecutive windows to overlap by $75\%$. 

We quantitatively evaluate the performance of estimating normals from sidescan intensity using the following metrics: mean absolute error (MAE), mean absolute relative error (REL), root mean squared error (RMSE) and accuracy under threshold ($\delta<1.05^i, i=1,2,3$), following our previous works \cite{xie2019inferring}\cite{xie2021}.

The CNN model is trained on a single Nvidia GEFORCE RTX 2080 Ti GPU. We train the model with a batch size of 64 with the Adam optimizer with parameters $(\beta_1=0.9, \beta_2=0.999)$ and a linear decayed learning rate with initial value $2\mathrm{e}{-4}$. Besides the normal-aware loss, we also use total variation regularization both across-track and along-track assuming some smoothness of the seafloor.

\subsection{Training SIREN}
For Dataset \RNum{1}, the test set for the CNN contains 12 survey lines of sidescan, half of which are perpendicular to the other half. These lines are also used to train the SIREN, the implicit representation of the bathymetry $\Phi_\theta$.  The SIREN network is initialized to random values.  During training, it is used to generate surface normals which are then compared to those generated by the CNN from the sidescan. Since we are using the sine function as the activation function in SIREN, the output of the SIREN is bounded between the range $[-1,1]$. To represent the bathymetry, we also scale all positional data so that the $x,y$ positions of the bathymetry are within the range $[-1,1]$. We use a 5-layer SIREN MLP with a hidden layer size of 128. We train the network for 1500 epochs, with an initial learning rate of $2\mathrm{e}{-4}$ that decays by a factor of 0.995 every epoch. Each batch contains 64 randomly sampled sidescan pings and 2000 randomly sampled altimeter points.  For each ping we use all the valid intensities from both port and starboard head (excluding the nadir and water column). We find that using all valid bins in the sidescan ping makes the SIREN training process less noisy, compared to randomly sampling 64 bins as in \cite{nils2021}. The training of SIREN requires approximately 12 hours on a single Nvidia GEFORCE RTX 2080 Ti GPU. We remark that the SIREN converges already pretty well in large scale at 800-1000 epochs, though much more epochs with smaller and smaller  learning rate is needed for the little details such as rocks.

For Dataset \RNum{2}, the whole dataset, which contains 36 survey lines of sidescan from freshwater, is used to train the SIREN for reconstructing the bathymetry of the surveyed lake. We keep most of the hyper-parameters the same as training SIREN for Dataset \RNum{1}, except that we use layers of half the width, trained with 1000 epochs due to the bathymetry from Dataset \RNum{2} being much less complex.

\section{Results}\label{par:results}
\begin{figure}[t]
\centering
\includegraphics[width=3.5in]{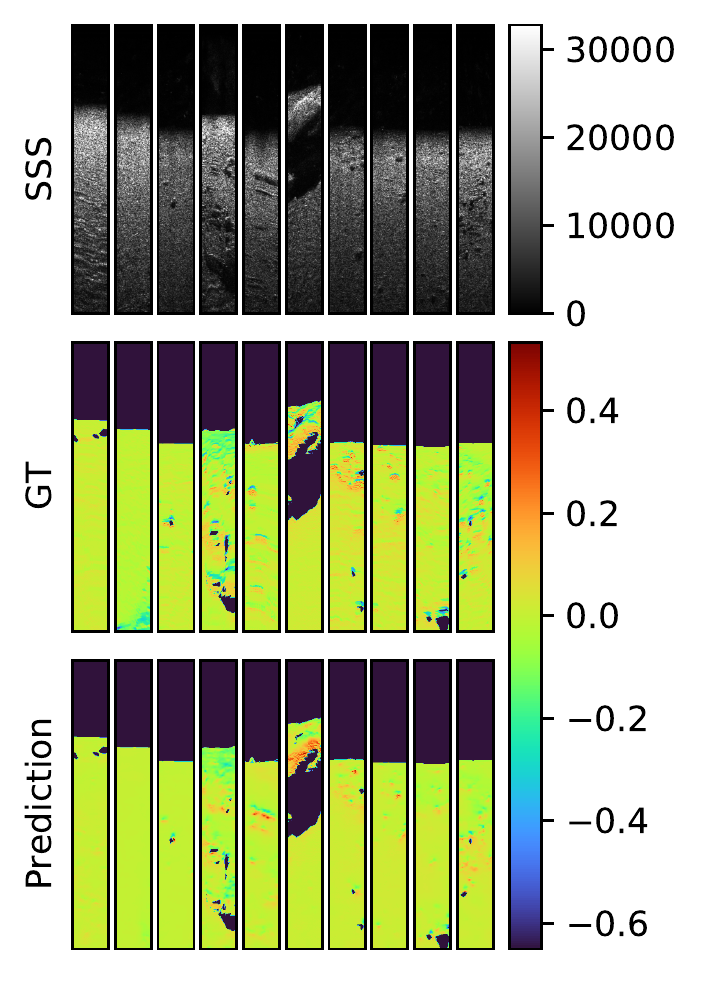} 
\caption{Here we show examples of the performance of the CNN in predicting surface normals in the plane perpendicular to the transducer array.  Top: the examples sidescan images fed to the CNN; Middle: the normals as computed from the MBES mesh, ground truth; Bottom: the normals predicted by the CNN; Note that the invalid points are masked (black) due to shadows or water columns.}
\label{fig:sss-gt-pre}
\end{figure}
\subsection{Normal Estimation with a CNN}
Initially we evaluate how well the CNN could reproduce the seabed surface normal. The CNN predictions after training are shown in Fig. \ref{fig:sss-gt-pre}, where one can see that the CNN has been able to learn to reproduce some of the rocks and smaller details but not all. The quantitative evaluation is shown in Table \ref{tab:quantitative-result-cnn}, where for error (the lower the better) we calculate mean absolute error (MAE), mean absolute relative error (REL), root mean squared error (RMSE) and for accuracy (the higher the better) we calculate the accuracy under threshold ($\delta<1.05^i, i=1,2,3$). We can see that the CNN generalizes pretty well on Dataset \RNum{2} despite that it comes from a different natural environment.
\begin{table*}[!t]
\renewcommand{\arraystretch}{1.5}
\caption{Quantitative Results For CNN}
\label{tab:quantitative-result-cnn}
\centering
\begin{tabular}{c||c|c|c|c|c}
\hline
& Error & Accuracy, \% \\

  & \multicolumn{1}{c|}{MAE \;\;\;\;REL\;\;\;\; RMSE} &  \multicolumn{1}{c|}{$\delta< 1.05$ \;$\delta< 1.05^2$ \;$\delta< 1.05^3$} \\
\hline

Dataset \RNum{1}-Testing & \multicolumn{1}{c|}{0.0221 \;0.0220\; 0.0297} & \multicolumn{1}{c|}
{92.10 \;\;\;\; 98.35 \;\;\;\; 99.49}\\
\hline
Dataset \RNum{2} &  \multicolumn{1}{c|}{0.0237 \; 0.0261 \;0.0341} & \multicolumn{1}{c|}
{91.70 \;\;\;\; 96.57 \;\;\;\; 97.97}\\

\hline
\end{tabular}
\end{table*}

\subsection{Bathymetric Mapping with SIREN Optimization}
\begin{figure*}[!t]
\centering
\includegraphics[width=6.5in]{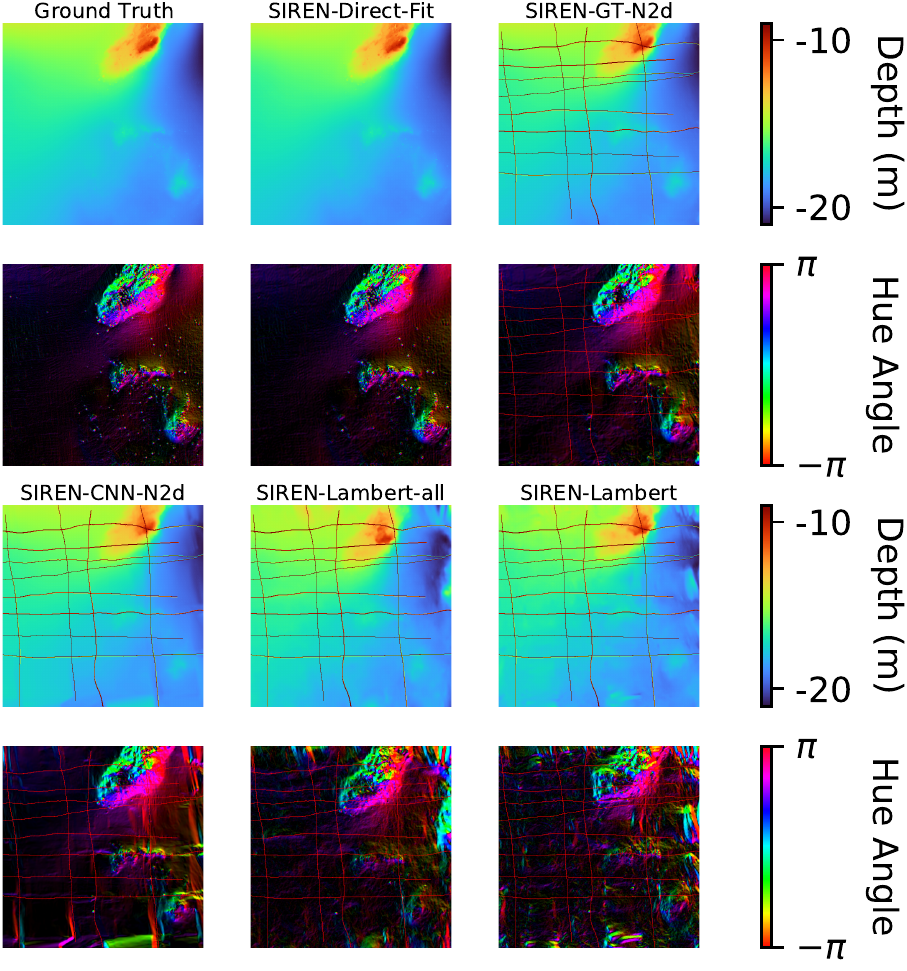} 
\caption{The results for the final bathymetry maps (approximately 250m$\times$250m) and their gradients in HSV, where hue is defined as the arctan(dx/dy), where saturation is set to 1 and value set to the norm of the gradient.  Row 1 (bathymetry) and 2 (gradient): MBES Ground Truth, the SIREN trained with the MBES, and the SIREN trained with the MBES normals projected into the sidescan planes; Row 3 and 4: The SIREN trained with CNN normals, the SIREN trained with Lambertian models with all sidescan lines, and the SIREN trained with Lambertian model with 12 lines. }
\label{fig:siren-cnn-all}
\end{figure*}

For Dataset \RNum{1}, we begin by confirming that the size of a SIREN is able to represent fully the complexity of the scene in the level of detail of the MBES ground truth.  We fit the grid-based multibeam bathymetry to a 5-layer SIREN with a hidden layer of size 128. In other words, we train the mapping function $\Phi_\theta: \mathbb{R}^2 \to \mathbb{R}$ to match the MBES bathymetric map directly. As we can see from Fig. \ref{fig:siren-cnn-all}, the MBES bathymetry (SIREN-Direct-Fit) gives us high-quality reconstructed bathymetry and its gradient, showing that the SIREN is capable of representing a scene of such complexity. 

In order to better see explicitly how different effects contribute errors to the final sidescan generated bathymetric map, we train the SIREN using the MBES again, but this time only using the information that ideally could be extracted from the sidescan pings we will use for mapping.    That is we train using the 2D normal projected to the sidescan sonar frame of the 12 survey lines.  As we mentioned at the end of Section \ref{sec:sidescan-draping}, for a sidescan ping, i.e., a single swath of sidescan, only the 2D normal projected to the sonar frame would affect the returned intensity, thus being encoded in the sidescan data. The information of the component of normal that is along the longitudinal axis cannot be recovered from a single observation. We train the SIREN with 12 lines of altimeter points and 12 lines of sidescan with "ground-truth" 2D projected normals computed from the MBES bathymetry, (SIREN-GT-N2d) in Fig. \ref{fig:siren-cnn-all}.  That can then be compared to the map from training the SIREN with 12 lines of altimeter points and their sidescan data with normals estimated from the CNN, (SIREN-CNN-N2d) in Fig. \ref{fig:siren-cnn-all}. 

\begin{figure*}[!ht]
\centering
\includegraphics[width=6.5in]{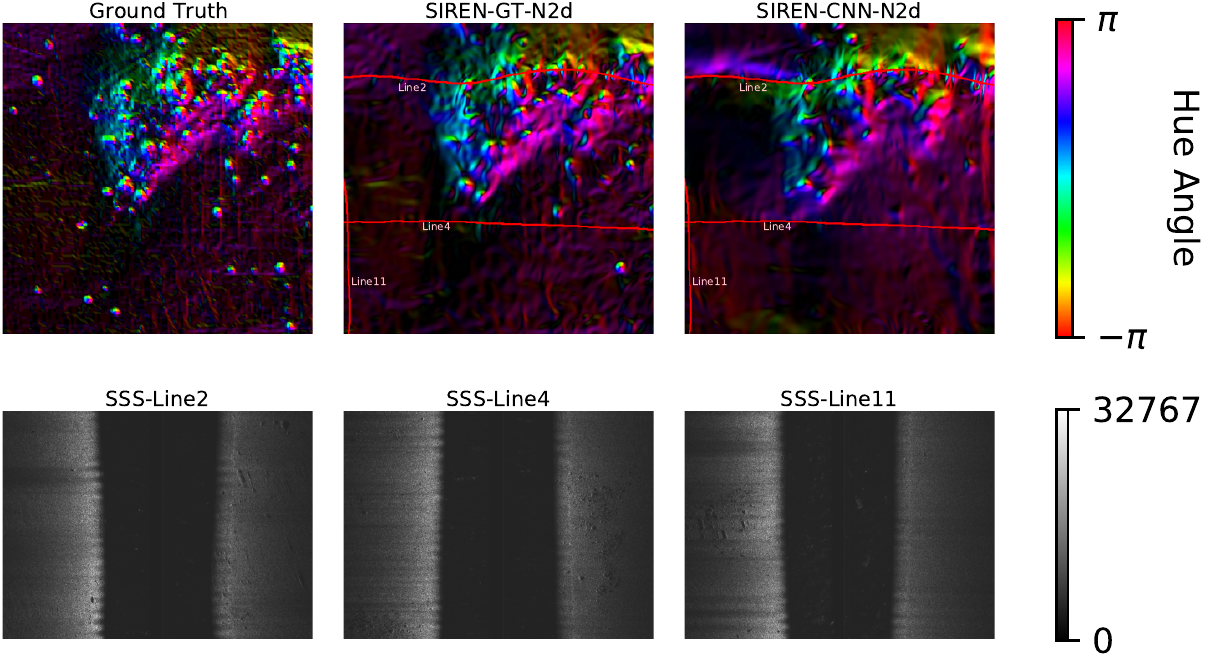} 
\caption{Zoomed-in areas including a ridge (approximately 50m$\times$50m). Top: gradient of the bathymetric map in HSV; left: ground truth gradient of our bathymetry from MBES data; middle: using normal computed from the MBES data to constrain the SIREN; right: using normal predicted from the CNN to constrain the SIREN. Bottom: corresponding SSS images from different lines.  }
\label{fig:siren-cnn-all-area6}
\end{figure*}

In SIREN-GT-N2d we observe that some detailed information, especially small rocks are missing compared to the gradient image from training directly on the MBES bathymetry. We argue that this is the upper-bound result of what we can be produced even if the CNN would estimate the 2D normals perfectly consistent with the MBES.  Notice that this worsening is completely dependant on the number of lines used.  If we were to, for example, double that to 24 lines, the map would look much closer to the MBES.
From SIREN-CNN-N2d in Fig. \ref{fig:siren-cnn-all} we see that we can reconstruct the large-scale structure, the hills and boulders, and even some of the smaller rocks with high quality, but some artifacts are introduced as can be seen easiest when we look at the gradient image. Note that the artifacts prominently appear at the peripheral due to lack of enough overlapping observations or additional constraints from the altimeter points to reduce the errors. 

\begin{figure*}[t]
\centering
\includegraphics[width=6.5in]{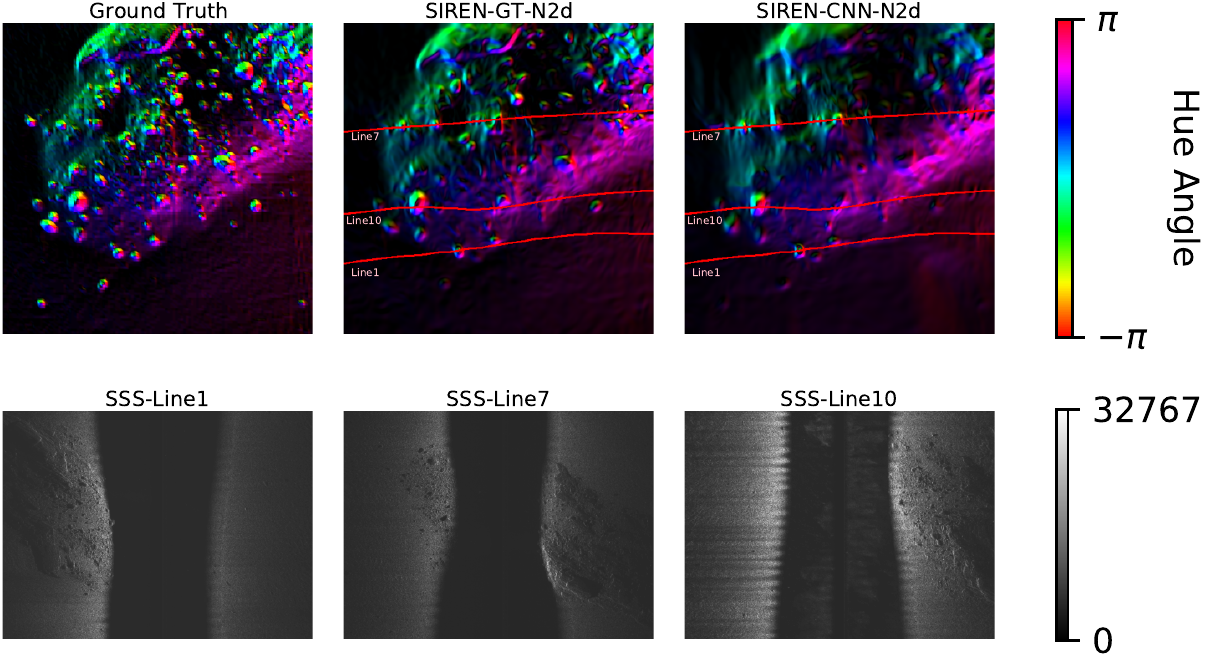} 
\caption{Zoomed-in areas including a hill (approximately 50m$\times$50m). Top: gradient of the bathymetric map in HSV; left: ground truth gradient of our bathymetry from MBES data; middle: using normal computed from the MBES data to constrain the SIREN; right: using normal predicted from the CNN to constrain the SIREN. Bottom: corresponding SSS images from different lines.}
\label{fig:siren-cnn-all-area7}
\end{figure*}

Fig. \ref{fig:siren-cnn-all-area6} and Fig. \ref{fig:siren-cnn-all-area7} show the zoomed-in part of the seafloor (gradient map) and the corresponding sidescan data for SIREN-GT-N2d and SIREN-CNN-N2d. From SIREN-GT-N2d and SIREN-CNN-N2d in Fig. \ref{fig:siren-cnn-all-area6} and \ref{fig:siren-cnn-all-area7} we can observe the reconstruction of small rocks by the proposed approach more clearly, especially between the sidescan lines. We can notice the degree of degradation between MBES and SIREN-GT-N2d and that between SIREN-GT-N2d and SIREN-CNN-N2d. In SIREN-CNN-N2d from Fig. \ref{fig:siren-cnn-all-area6} and \ref{fig:siren-cnn-all-area7}, however, we can still notice that some small rocks are reproduced from sidescan even though a few of them are not as sharp as in MBES. But the fact that some of the small rocks are clearly identified indicates that the proposed method is valuable for high-resolution bathymetry mapping with sidescan sonar.

We also ran the experiments with the modeling approach based on Lambertian assumption in \cite{nils2021} to compare the model-based approach and the data-driven approach. SIREN-Lambert-all in Fig. \ref{fig:siren-cnn-all} represents the experiment with the same 12 lines of altimeter points, but \textit{all} the sidescan available including the training and test set and SIREN-Lambert in Fig. \ref{fig:siren-cnn-all} only uses sidescan and altimeter points from the test set, 12 lines. SIREN-Lambert-all in Fig. \ref{fig:siren-cnn-all} shows that when using all sidescan data to jointly optimize the physical model and bathymetry we can produce a bathymetry qualitatively comparable to SIREN-CNN-N2d, but when we use much fewer lines of sidescan, the quality of reconstructed bathymetry has a notable degradation.

\begin{figure*}[t]
\centering
\includegraphics[width=6.5in]{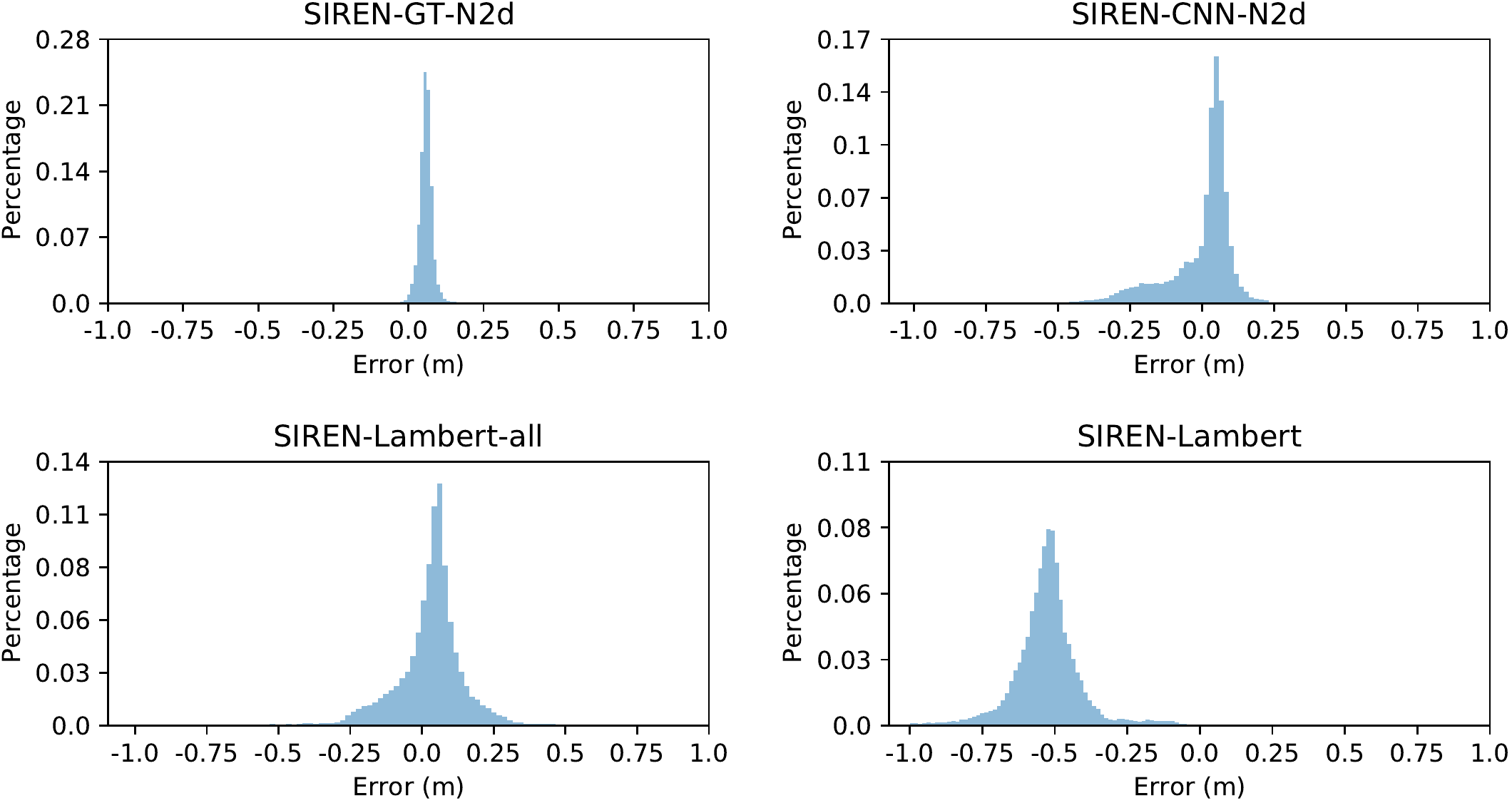} 
\caption{PDF curves of errors of the surveyed area. Top: the SIREN trained with the MBES normals projected into the sidescan planes (left) and the SIREN trained with CNN normals (right). Bottom: the SIREN trained with Lambertian models with all sidescan lines (left) and the SIREN trained with Lambertian model with 12 lines (right). }
\label{fig:pdf_curves}
\end{figure*}

For quantitative evaluation, we calculate the mean absolute error, \emph{MAE}, and the standard deviation, \emph{STD} of the signed errors on the bathymetry map and cosine similarity \emph{CS} on the gradient of the map (see Table \ref{tab:quantitative-result-siren}). 
We can notice from the table that all models have low errors on the bathymetry maps at the level of centimeters.  Also we can see that the MAE for both methods is the same and only $7\%$ of the error is due to the sonar model.  Most of the error is due to the effect of sampling the gradient projected onto the vertical planes of the sonar beam and the SIREN optimization. However, in terms of the cosine similarity score on the gradient map, data-driven approaches have considerably higher scores than model-based approaches, possibly due to data-driven approaches producing fewer artifacts on the bathymetry map. Furthermore, SIREN-CNN-N2d has achieved comparable performance with SIREN-GT-N2d, which is considered to be the upper-bound result.  
We also plot the probability distribution function (PDF) curves of the signed errors on the bathymetry, as shown in Fig. \ref{fig:pdf_curves}. We can visually see that the PDF curve for SIREN-GT-N2d has the minimum spread which is also reflected on the STD metric in Table \ref{tab:quantitative-result-siren}. Compared to SIREN-CNN-N2d, the spread of the PDF of for SIREN-Lambert-all is slightly larger. As for SIREN-Lambert, there is a noticeable bias in the error (around $-0.6$ m) and it has the largest spread of the PDF curve.

\begin{table}[!t]
\renewcommand{\arraystretch}{1.5}
\caption{Quantitative Results For SIREN on Dataset \RNum{1}}
\label{tab:quantitative-result-siren}
\centering
\begin{tabular}{c||c|c|c}
\hline
 & MAE (m) & STD ($\pm$m) & Gradient CS \\
\hline
SIREN-GT-N2d & 0.079 & 0.025 & 0.740\\
\hline
SIREN-CNN-N2d & 0.085 & 0.163 & 0.738\\
\hline
SIREN-Lambert-all & 0.077 & 0.140 & 0.529\\
\hline
SIREN-Lambert & 0.085 & 0.174 & 0.339\\
\hline

\end{tabular}
\end{table}

\begin{figure*}[!t]
\centering
\includegraphics[width=6.25in]{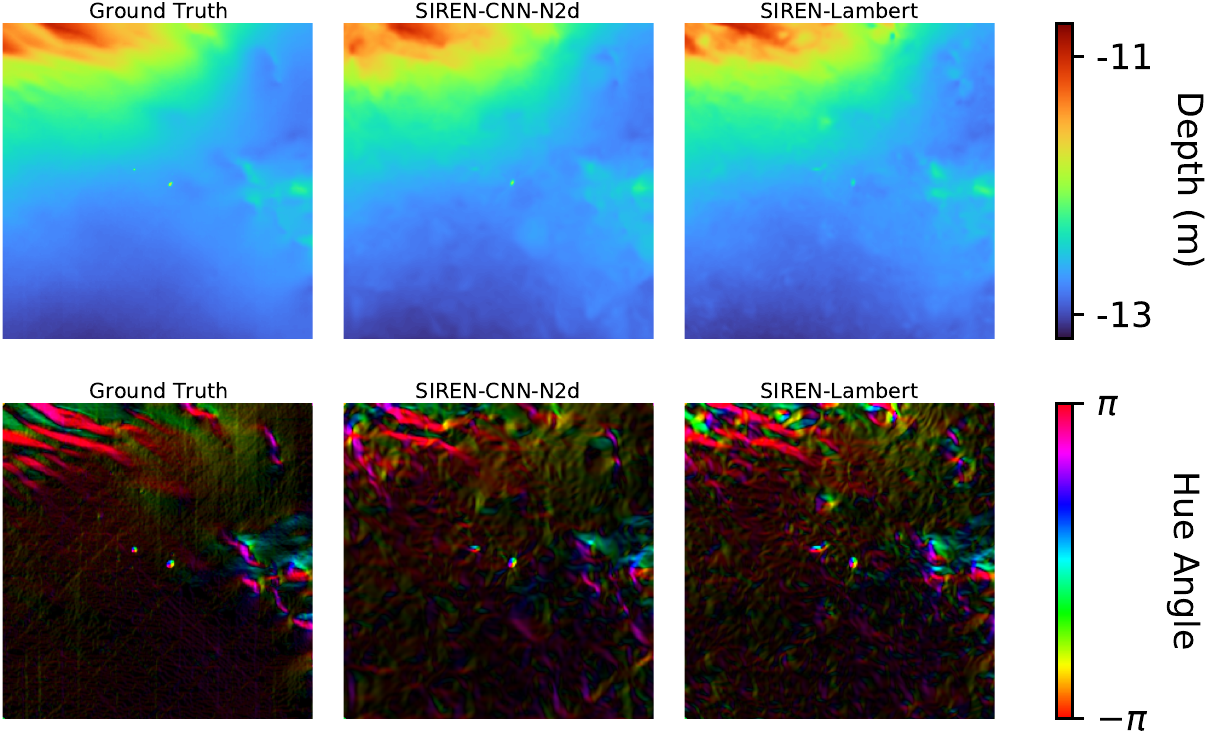} 
\caption{The results for the final bathymetry maps from Dataset \RNum{2}, another place in freshwater.  Top: bathymetry maps (approximately 300m$\times$300m);  bottom: gradient. Column 1, 2 and 3: MBES Ground Truth, the SIREN trained with CNN normals from all the sidescan lines, the SIREN trained with Lambertian models with all the sidescan lines.}
\label{fig:motala-cnn-lambert}
\end{figure*}

In order to test the generalization ability of the CNN on a different environment, we train a SIREN from Dataset \RNum{2}, where the CNN has not been trained on and compare the results with the data collected with a MBES (see Fig.  \ref{fig:motala-cnn-lambert}). We use all available sidescan data and altimeter data from the whole survey to train the SIREN. We also compare with a model-approach method using all the data from Dataset \RNum{2}. The quantitative evaluation is in Table  \ref{tab:quantitative-result-siren-dataset2}.   From SIREN-CNN-N2d in Fig.  \ref{fig:motala-cnn-lambert} we can see that the proposed method could reproduce the large-scale features of the bathymetry, showing the generalization ability of the CNN to a different environment in shallow water. However, we can also see that from both SIREN-CNN-N2d and SIREN-Lambert, some artifacts are introduced, as can be seen easiest in the gradient image. Qualitatively it is difficult to tell which map is better, but the quantitative result in Table \ref{tab:quantitative-result-siren-dataset2} again shows that using a data-driven approach to model the sidescan has a  higher cosine similarity on the gradient.

\begin{table}[!t]
\renewcommand{\arraystretch}{1.5}
\caption{Quantitative Results For SIREN on Dataset \RNum{2}}
\label{tab:quantitative-result-siren-dataset2}
\centering
\begin{tabular}{c||c|c|c}
\hline
 & MAE (m) & STD ($\pm$m)  & Gradient CS \\
\hline
SIREN-CNN-N2d & 0.018 & 0.049 & 0.591\\
\hline
SIREN-Lambert & 0.016 & 0.060 &0.513\\
\hline
\end{tabular}
\end{table}
\section{Discussion}\label{par:discussion}
  It is apparent by looking at the fifth sidescan image  in Fig. \ref{fig:sss-gt-pre} that there is a significant ridge feature near the nadir that one could easily believe is better reproduced by the CNN than the MBES.  This highlights the difficulty in training this network as the sidescan may show some features more clearly than the MBES, which is one of the reasons both are often used together. We, however, are forced to treat the MBES as ground truth in this image and consider the bottom as an imperfect prediction. 

 From the comparison between MBES, SIREN-GT-N2d and SIREN-CNN-N2d in Fig. \ref{fig:siren-cnn-all-area6}, Fig. \ref{fig:siren-cnn-all-area7} and Table \ref{tab:quantitative-result-siren}, we can hypothesize that we lose more information in the process of 2D normal projection than the CNN's prediction since the degradation of the map is more noticeable between MBES and SIREN-GT-N2d, both qualitatively and quantitatively. However, as aforementioned, the worsening due to 2D normal projection can be alleviated by increasing the number of lines we used to survey the area. It is a trade-off between the quality of the map and the efficiency of the survey.

From the comparison between SIREN-Lambert-all, SIREN-Lambert and SIREN-CNN-N2d in Fig. \ref{fig:siren-cnn-all} and the cosine similarity in Tables \ref{tab:quantitative-result-siren} and \ref{tab:quantitative-result-siren-dataset2}, we can hypothesize that the Lambertian model is not as good at modeling the sidescan reflection process as the trained CNN when it only has the 12 lines to estimate with.  Also that using more than 12 lines to optimize results in better final maps.  This is not surprising since the Lambertian based approach includes learning of the reflection process parameters such as beam width, reflectivity and time-varying gain, all of which the CNN model learned separately on other data. However, the considerable advantage of using the Lambertian model is that no ground truth bathymetry is required for optimization. This means that the comparisons here (evaluated on Dataset \RNum{1}) are not quite fair between the two approaches. Nevertheless, if we consider that the CNN might just as well have been trained on a similar bottom with the same sensor setup and then we only had the 12 lines from the test set of this area, it would have been a fair comparison.  As it stands the CNN did not need to generalize to a new area even if it did not use the test lines for training, which is the case when we assess the proposed method on Dataset \RNum{2}.

That the cosine similarity on gradient map with data-driven approaches is noticeably higher than that model-based approaches, on both Dataset \RNum{1} and Dataset \RNum{2} helps to support the hypothesis that CNN performs better at the sidescan reflection process than a simple Lambertian model.  That Dataset \RNum{2} is not seen by the CNN eliminates the possibility of CNN overfitting area-specific artifacts in the MBES mesh. However, the cosine similarity with data-driven approaches on Dataset \RNum{1} being much higher is in part due to the fact that CNN has been trained on the same seafloor.

The experiment run on Dataset \RNum{2} shows one application of the proposed method, that is, reconstructing survey-scale bathymetry from another place with the same ship and equipment once the CNN is trained in one place.   This then shows that once trained the approach will work with only a sidescan sonar.  Both methods however require accurate navigation to give these results.  For an AUV the dead-reckoning will degrade the results but we did not investigate that effect here.  

\subsection{Limitation}
All bathymetric surveys require accurate navigation estimates but the method presented here is even more sensitive to this as we fuse data from several lines of the survey.  Any drift in the navigation estimates between lines will impact the results.  To use this on an AUV some mechanism to control drift in dead-reckoning must be used or some post-processing to correct the navigation be applied. 

The time for training a SIREN, which usually costs approximately 12 hours on a modern desktop, does limit the real-time application of this approach. As a common problem in the community of Implicit Representation Learning, there is some very recent research \cite{Garbin_2021_ICCV} addressing such an issue. Furthermore, the proposed method and the Lambertian model both lack the ability to model the shadows in the sidescan images. In the proposed method, shadows are masked out as invalid data; thus the surface normals there do not contribute to the SIREN estimation; in the Lambertian model, a fixed threshold is used to remove the shadows. Another limitation is that we are forced to treat the MBES as the ground truth, which limits the resolution of reconstructed bathymetry despite the high across-track resolution of the sidescan. 

\section{Conclusion}\label{par:Conclusion}
We present a data-driven approach that uses sidescan and sparse altimeter readings to reconstruct a continuous bathymetric map represented by a neural network. We demonstrate that such an approach is capable of efficiently building a high-quality bathymetry with only a few sidescan survey lines. And compare this to a Lambertian model-based approach using the same optimization framework.

One can conclude that both approaches work given enough data.  The CNN approach can be trained separately to learn the model and the data need not be from the exact same area, while the Lambertian model does not need any MBES to learn but requires more data from the actual survey area for the same results. The proposed method has an advantage when we cannot collect too much data or when we want to cover a wider area.

\section{Future work}\label{par:Future_work}
One possible future work is to combine the MBES with raw sidescan data without downsampling together to produce a super-resolution bathymetry, fully utilizing the advantages of both sensors. We would need to replace the CNN with, for example, a differentiable renderer for the sidescan modeling. Another possible future direction to do is to jointly optimize the sidescan sonar's poses with bathymetry since the attitude accuracy of the RTK GPS is not sufficient when modeling the raw sidescan data. Because small attitude errors can result in large deviation of estimating the intersection between sidescan and seafloor, especially as one moves further away from the sidescan sensor within one sidescan ping. One can start investigating them using simulated datasets, which contain higher resolution ground truth bathymetry and ground truth poses. 

\section*{Acknowledgment}
This work was partially supported by the Wallenberg AI, Autonomous
Systems and Software Program (WASP) and  by the Stiftelsen  för  Strategisk  Forskning (SSF)  through  the  Swedish  Maritime  Robotics  Centre  (SMaRC)(IRC15-0046). Our dataset was acquired in collaboration with MarinMätteknik (MMT) Gothenburg.

\ifCLASSOPTIONcaptionsoff
  \newpage
\fi

\bibliographystyle{IEEEtran}
\bibliography{IEEEabrv,ieeeref}



%

\begin{IEEEbiography}[{\includegraphics[width=1in,height=1.25in,clip,keepaspectratio]{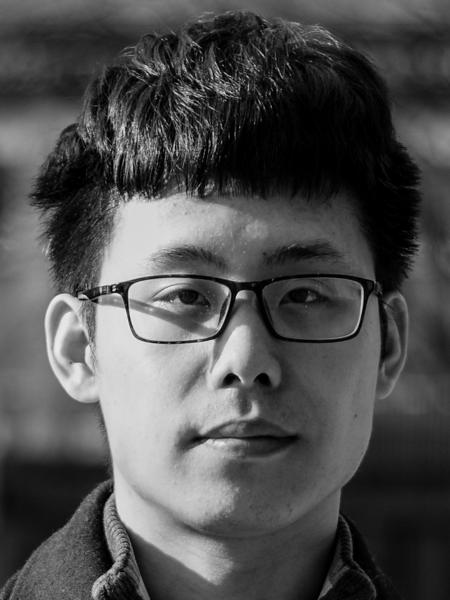}}]{Yiping Xie}
received the B.S. degree in electrical engineering from Beihang University, Beijing, China, in 2017, and the M.Sc. degree in computer science from Royal Institute of Technology (KTH), Stockholm, Sweden, in 2019. 

He is currently a Ph.D. student with the Wallenberg AI, Autonomous Systems and Software Program (WASP) from the Robotics Perception and Learning Lab at KTH. His research interests include perception for underwater robots, bathymetric mapping and localization with sidescan sonar.
\end{IEEEbiography}

\begin{IEEEbiography}[{\includegraphics[width=1in,height=1.25in,clip,keepaspectratio]{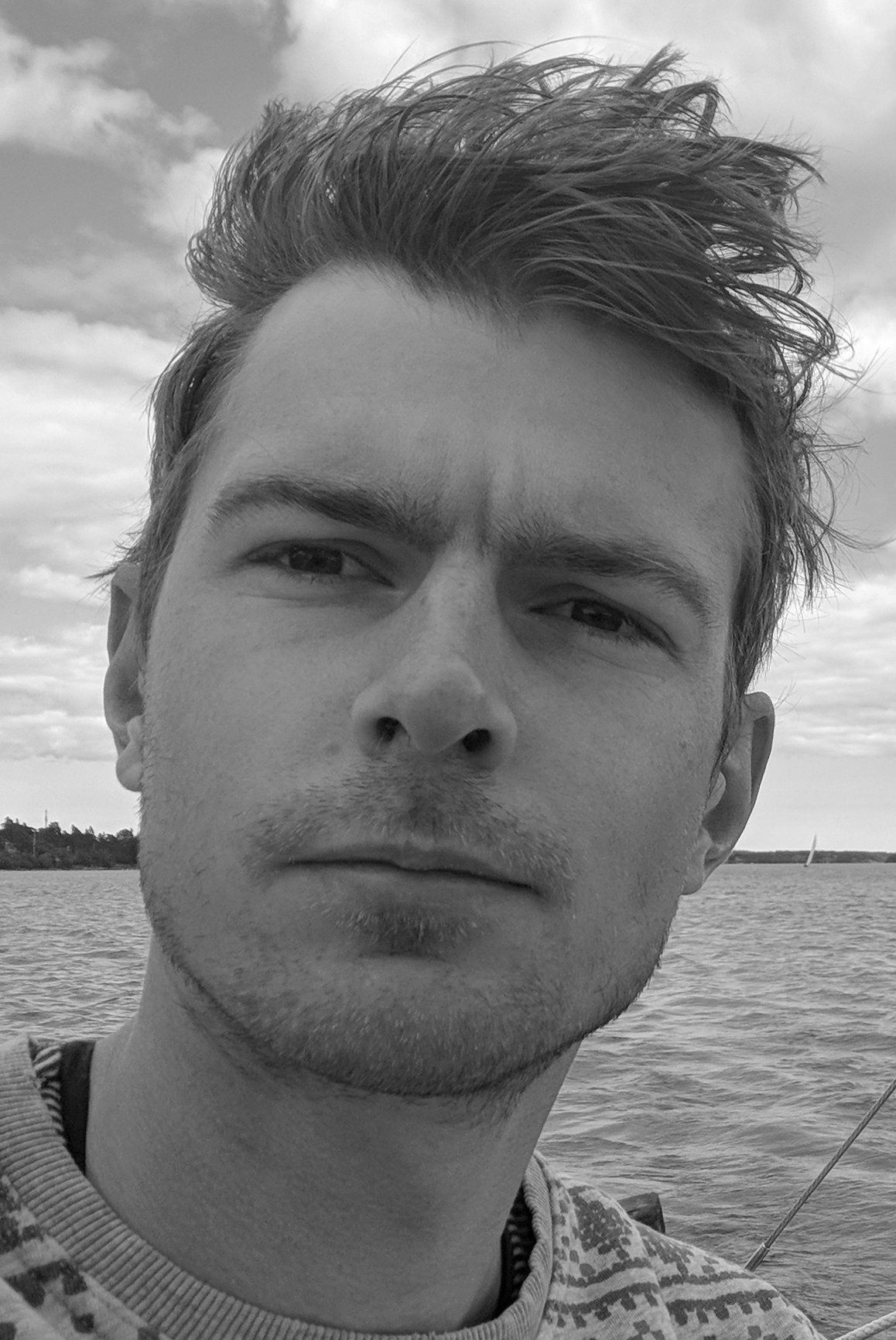}}]{Nils Bore} received the M.Sc. degree in mathematical engineering from the Faculty of Engineering, Lund University, Lund, Sweden, in 2012, and the Ph.D. degree in computer vision and robotics from the Robotics Perception and Learning Lab, Royal Institute of Technology (KTH), Stockholm, Sweden, in 2018.
He is currently a researcher with the Swedish Maritime Robotics (SMaRC) project at KTH. His research interests include robotic sensing and mapping, with a focus on probabilistic reasoning and inference. Most of his recent work has been on applications of specialized neural networks to underwater sonar data. In addition, he is interested in system integration for robust and long-term robotic deployments.
\end{IEEEbiography}


\begin{IEEEbiography}[{\includegraphics[width=1in,height=1.25in,clip,keepaspectratio]{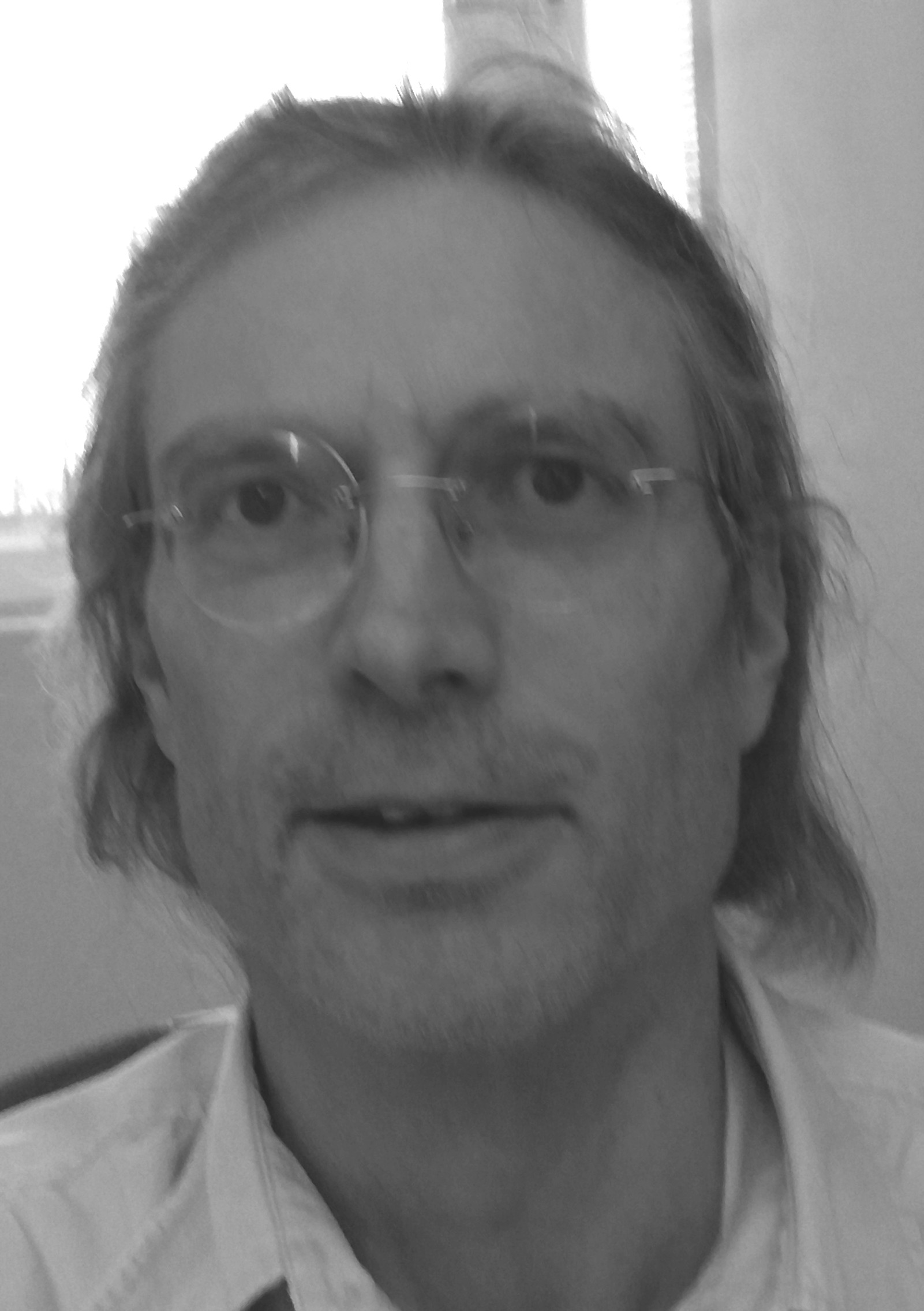}}]{John Folkesson} received the B.A. degree in physics from Queens College, City University of New York,
New York, NY, USA, in 1983, and the M.Sc. degree in computer science, and the Ph.D. degree in robotics
from Royal Institute of Technology (KTH), Stockholm, Sweden, in 2001 and 2006, respectively.
He is currently an Associate Professor of robotics with the Robotics, Perception and Learning Lab, Center for Autonomous Systems, KTH. His research interests include navigation, mapping, perception, and
situation awareness for autonomous  robots.
\end{IEEEbiography}




\end{document}